\documentclass[runningheads]{llncs}

\usepackage{eccv}
\usepackage{eccvabbrv}

\usepackage{graphicx}
\usepackage{booktabs}
\usepackage{color}
\usepackage{xcolor,colortbl}
\usepackage{pifont}
\usepackage{bbding}
\usepackage{multirow}
\usepackage{subcaption}
\usepackage{makecell}
\usepackage{titletoc}
\usepackage{etoolbox}
\usepackage[accsupp]{axessibility}  % Improves PDF readability for those with disabilities.

\usepackage[pagebackref,breaklinks,colorlinks,citecolor=eccvblue]{hyperref}

\usepackage{orcidlink}

\usepackage[lined,boxed,commentsnumbered]{algorithm2e}
\usepackage{caption}

\newcommand{\model}{SparseAD\xspace}
\begin{document}
\bibliographystyle{unsrt}
\title{\model: Sparse Query-Centric Paradigm for Efficient End-to-End Autonomous Driving} 
\titlerunning{\model, End-to-End Autonomous Driving Method}
\author{
Diankun Zhang$^{2,\star,\dagger}$\and
Guoan Wang$^{1,\star}$ \and
Runwen Zhu$^{3,\star,\dagger}$ \and
Jianbo Zhao$^{4,\star,\dagger}$ \and \\
Xiwu Chen$^{1}$ \and
Siyu Zhang$^{1}$\and
Jiahao Gong$^{1}$ \and
Qibin Zhou$^{1}$ \and
Wenyuan Zhang$^{1}$ \and \\
Ningzi Wang$^{1}$ \and 
Feiyang Tan$^{1,\ddag}$ \and
Hangning Zhou$^{1,\ddag}$ \and
Ziyao Xu$^{1}$ \and
Haotian Yao$^{1}$ \and \\
Chi Zhang$^{1}$ \and
Xiaojun Liu$^{2}$ \and
Xiaoguang Di$^{3}$ \and
Bin Li $^{4}$
\\
}
\authorrunning{Zhang, Wang, Zhu, Zhao, et al.}
\institute{$^{\text{1}}$Mach Drive 
$^{\text{2}}$University of Chinese Academy of Sciences \\
$^{\text{3}}$Harbin Institute of Technology
$^{\text{4}}$University of Science and Technology of China 
}

\maketitle
{\let\thefootnote\relax\footnote{{\text{$^{\star}$} denotes equal contributions.}}}
{\let\thefootnote\relax\footnote{{\text{$^{\dagger}$} denotes this work was done during the internship at Mach Drive.}}}
{\let\thefootnote\relax\footnote{\text{$^\ddag$} denotes the corresponding author.}}
\begin{abstract}

End-to-End paradigms use a unified framework to implement multi-tasks in an autonomous driving system. Despite simplicity and clarity, the performance of end-to-end autonomous driving methods on sub-tasks is still far behind the single-task methods. Meanwhile, the widely used dense BEV features in previous end-to-end methods make it costly to extend to more modalities or tasks.
In this paper, we propose a \textbf{Sparse} query-centric paradigm for end-to-end \textbf{A}utonomous \textbf{D}riving (\textbf{SparseAD}), where the sparse queries completely represent the whole driving scenario across space, time and tasks without any dense BEV representation.
Concretely, we design a unified sparse architecture for perception tasks including detection, tracking, and online mapping. Moreover, we revisit motion prediction and planning, and devise a more justifiable motion planner framework. On the challenging nuScenes dataset, SparseAD achieves SOTA full-task performance among end-to-end methods and significantly narrows the performance gap between end-to-end paradigms and single-task methods. Codes will be released soon.

    
\end{abstract}
 \section{Introduction}

\begin{figure}[tb]
  \centering
  \begin{subfigure}{0.6\linewidth}
  \begin{subfigure}{1\linewidth}
    \includegraphics[width=1\linewidth]{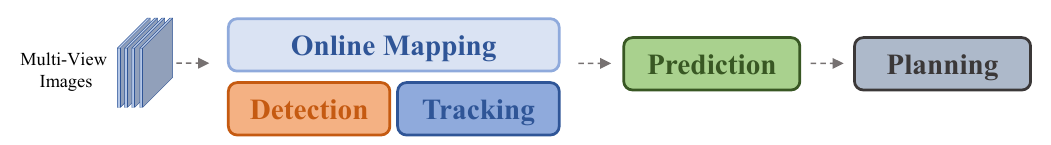}
    \caption{Traditional autonomous driving paradigm}
    \label{fig:pipeline-a}
  \end{subfigure}
  \hfill
  \\
  \begin{subfigure}{1\linewidth}
    \includegraphics[width=1\linewidth]{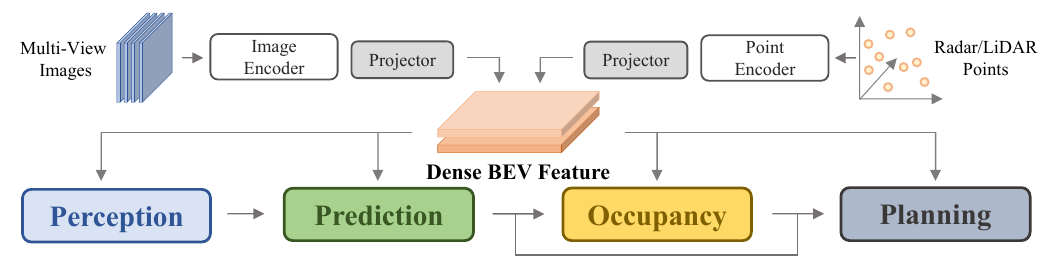}
    \caption{Dense BEV-Centric end-to-end paradigm}
    \label{fig:pipeline-b}
  \end{subfigure}
  \label{fig:short}
  \\
    \begin{subfigure}{1\linewidth}
    \includegraphics[width=1\linewidth]{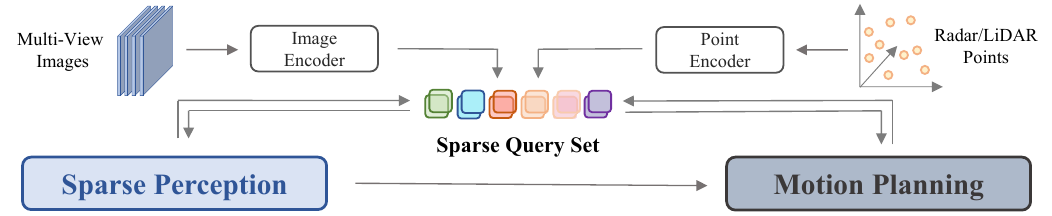}
    \caption{Sparse Query-Centric end-to-end paradigm}
    \label{fig:pipeline-c}
  \end{subfigure}
  \hfill
  \end{subfigure}
  \begin{subfigure}{0.38\linewidth}
  \centering
    \begin{subfigure}{1\linewidth}
    \includegraphics[width=1\linewidth]{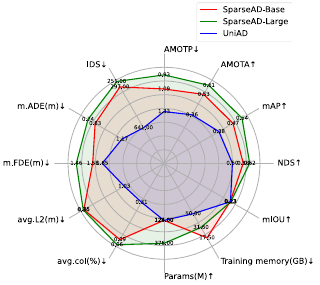}
    \caption{The comprehensive performance comparison between the proposed method with state-of-the-arts end-to-end autonomous driving method\cite{hu2023planning}}
    \label{fig:pipeline-d}
    \end{subfigure}
    \\
    \begin{subfigure}{1\linewidth}
    \end{subfigure}
    \\
    \begin{subfigure}{1\linewidth}
    \end{subfigure}

  \end{subfigure}
  
  \caption{The comparison of mainstream autonomous driving paradigms. (a) The traditional paradigm. (b) The Dense BEV-Centric paradigm. (c) The proposed Sparse Query-Centric end-to-end autonomous driving paradigm. And (d) the comprehensive performance comparison between the representative method of (b) and (c).}
  \label{fig:pipeline}
\end{figure}

In recent years, deep learning and AI technologies have become increasingly vital in autonomous driving systems, which require correct decisions in complex driving scenarios to ensure the driving safety and comfort. Generally, an autonomous driving system gathers multiple tasks such as detection, tracking, online mapping, motion prediction, and planning. 
As shown in \cref{fig:pipeline-a}, traditional modular paradigm decouples the complicated system into several single tasks which are optimized independently. In this paradigm, hand-crafted post-processing is set between standalone single-task modules, which makes the whole pipeline much more bloated. On the other hand, with the lossy compression of scenario information between stacked tasks, the error of entire system accumulates gradually, which may lead to potential safety problems \cite{liang2020pnpnet, sadat2020perceive}.

Regarding the issues above, end-to-end autonomous driving systems take raw sensor data as input and return planning results in a more succinct way. Earlier works propose to skip intermediate tasks and predict the planning results directly from raw sensor data \cite{codevilla2018end,codevilla2019exploring, wu2022trajectory, zhang2021end}. While much more diametrical, it is unsatisfactory in terms of model optimization, explainability, and planning performance. Another multifaceted paradigm with better interpretability is to integrate multiple parts of autonomous driving into a modular end-to-end model \cite{hu2023planning, ye2023fusionad}, in which multi-dimensional supervisions are introduced to improve the understanding of complex driving scenarios and brings the multi-task capabilities.

As shown in \cref{fig:pipeline-b}, in most previous modular end-to-end methods, the whole driving scenario is represented through dense Bird's Eye View (BEV) features including multi-sensor and temporal information, which are used as source input for full-stack driving tasks including perception, prediction and planning. Based on the fact that dense BEV feature does play a critical role in multi-modality and multi-tasks, across space and time, we summarize previous end-to-end methods with BEV representations into the paradigm of \textbf{Dense BEV-Centric}.
However, despite the conciseness and interpretability of the methods\cite{hu2023planning,ye2023fusionad}, their performance on each sub-task of autonomous driving still lags far behind the corresponding single-task methods\cite {liu2022petr, wang2023exploring, pang2023standing, jiang2023vad}. Furthermore, long-term temporal and multi-modality fusion are achieved mostly by multiple BEV feature maps under the dense BEV-Centric paradigm, which brings significant increase in computation cost, memory footprint, and more burden on real-world deployment \cite{zhang2023fully, jiang2023far3d}.

In this paper, we propose a novel \textbf{Sparse Query-Centric} paradigm for end-to-end Autonomous Driving (SparseAD), in which all elements across space and time in the whole driving scenario are represented by sparse queries without any dense BEV features, as shown in \cref{fig:pipeline-c}. With the sparse representation, it's much more efficient for end-to-end models to take advantage of longer historical information, extend to more modalities and tasks with conspicuously less computation cost and memory footprint.

Specifically, we revise the modular end-to-end architecture and simplify it into a succinct structure consisting of sparse perception and motion planner. In \textbf{Sparse Perception}, perception tasks including detection, tracking and online mapping are unified with general temporal decoders\cite{wang2023exploring}, in which multi-sensor features and historical memories are regarded as tokens, and object queries and map queries represent obstacles and road elements in the driving scenario respectively. In \textbf{Motion Planner}, with sparse perception queries as environmental representation, we apply multi-modal motion prediction for both ego-vehicle and surrounding agents simultaneously to get various initial solutions for ego-vehicle, then driving constraints of multiple dimensions are fully considered to generate final planning. It is noteworthy that the excellent efficiency of SparseAD makes it feasible to benefit from scaling up the model with advanced backbone networks \cite{dosovitskiy2020image,liu2022convnet}, large-scale pretrain \cite{fang2023eva,radford2021learning}, or massive data. We conduct substantial experiments on the popular nuSences dataset, which demonstrate the effectiveness and superiority of the proposed sparse query-centric paradigm for end-to-end autonomous driving as shown in \cref{fig:pipeline-d}.

In summary, the main contributions of this paper include:
\begin{itemize}

\item We propose a novel \textbf{Sparse Query-Centric} paradigm for end-to-end Autonomous Driving (\model) without any dense BEV representation, which has great potential to extend to more modalities and tasks efficiently.

\item We simplify the modular end-to-end architecture into sparse perception and motion planning, where the former unifies detection, tracking and online mapping in a totally sparse manner while the later applies motion prediction and planning in a more justifiable framework.

\item On the challenging nuScenes dataset\cite{caesar2020nuscenes}, \model achieves SOTA performance among end-to-end methods and significantly narrows the gap of performance between end-to-end paradigms and single-task methods, which demonstrates the great potential of the proposed sparse end-to-end paradigm.
\end{itemize}
\section{Related Work}
\label{sec:related}

\subsection{Detection and Tracking}
\label{sec:related-det-mot}

In recent years, 3D detection has been widely studied in many works, e.g., lidar-based methods \cite{yin2021center, chen2022focal, chen2023largekernel3d, chen2023voxelnext, zhang2023fully}, camera-based methods \cite{huang2021bevdet, li2022bevformer, liu2022petr, li2023bevdepth, wang2023exploring} and multi-modal fusion methods \cite{bai2022transfusion, liu2023bevfusion, yan2023cross}. We mainly introduce the camera-based methods since the proposed method focuses on this modal mostly. Due to the property of dense representation, the Bird's Eye View (BEV) features are widely used in many superior works. LSS \cite{philion2020lift} first proposes to project image features onto BEV features by depth prediction, based on which series follow-up works \cite{huang2021bevdet, li2023bevdepth, li2022bevformer, yang2023bevformer, han2023exploring, li2023bevstereo} achieved better detection performance. Another branch of 3D perception adopts the sparse route without any dense representation. DETR3D \cite{wang2022detr3d} first proposes to use 3D queries to aggregate multi-view 2D features to achieve 3D perception in a sparse manner. PETR and Sparse4D series \cite{liu2022petr, liu2023petrv2, wang2023focal, yan2023cross, zhang2023fully, lin2022sparse4d, lin2023sparse4dv2, lin2023sparse4d} further continuously explore efficient spatial-temporal aggregation solutions based on sparse queries entirely, which show significant advantages both in efficiency and performance.

Generally, tracking generates obstacle trajectories from adjacent multi-frame perception results for downstream tasks. Traditional algorithms \cite{weng20203d, yin2021center, wang2023exploring} adopt the "tracking-by-detection" paradigm and mainly focus on establishing the association between the tracked trajectories and new-come perception results, in which the performance of tracking is significantly affected by detection. Inspired by DETR \cite{codevilla2018end}, more works \cite{lin2021global, zeng2022motr, zhang2023motrv2, yu2023motrv3, sun2020transtrack, pang2022simpletrack, meinhardt2022trackformer} explore the paradigm of joint detection and tracking by propagating the queries and supervision associations between adjacent frames. MUTR3D \cite{zhang2022mutr3d} migrates this paradigm from 2D to 3D. PF-Track \cite{pang2023standing} and Sparse4Dv3 \cite{lin2023sparse4d} further emphasize and exploit spatial-temporal information effectively which improves the performance significantly. However, there is an apparent imbalance in the optimization of different kinds of queries during joint training \cite{yu2023motrv3}, which may markedly impair the performance of detection.

\subsection{Online Mapping}
\label{sec:related-mapping}

In autonomous driving, maps could provide rich scenario information which is crucial for driving safety. Previous solutions based on offline High-Definition maps (HD-Map) require a vast amount of human efforts and resources which limits its scalability \cite{li2022hdmapnet}. Recently, online mapping to construct HD-Map with only onboard sensors has drawn extensive attention \cite{li2022hdmapnet, liu2023vectormapnet, qiao2023end, ding2023pivotnet, yuan2024streammapnet}. HDMapNet \cite{li2022hdmapnet} predicts the maps by pixel-wise segmentation and heuristic post-processing. VectorMapNet \cite{liu2023vectormapnet} proposes to refine map elements subsequently by an auto-regressive transformer. BeMapNet \cite{qiao2023end} constructs instance-level vectorized HD-Map in an end-to-end manner and adopts piece-wise Bezier curves to model the map element. PivotNet \cite{ding2023pivotnet} further proposes a map construction method based on pivot-based representations. StreamMapNet \cite{yuan2024streammapnet} adopts a streaming strategy to gain performance enhancement from temporal fusion. However, all above methods use dense BEV features for environment representation whose computational cost grows squarely regarding to the perception distance, making it much more expensive to work together with other modules in the end-to-end models.

\subsection{Motion Prediction}
\label{sec:related-forecasting}

In traditional autonomous driving systems, motion prediction modules predict the future trajectories or intentions of surrounding agents to facilitate the correct decision-making and safe driving of ego-vehicle. Early methods render the driving scenario into rasterized images and predict future motions by off-the-shelf convolution networks\cite{chai2019multipath, phan2020covernet}. VectorNet \cite{gao2020vectornet} first proposes efficient vectorized representations for motion prediction and lays the foundation for many recent methods. Graph-based methods are also widely used to model and extract the relative relations across agents and maps \cite{zhao2021tnt, varadarajan2022multipath++, zhou2022hivt}. QCNet\cite{zhou2023query}, MTR series \cite{shi2022motion, shi2024mtr++} further explore the environmental representation in more efficient ways and improve prediction performance. PnPNet \cite{liang2020pnpnet} and ViP3D \cite{gu2023vip3d} put a lot of attempts and effort into applying trajectory prediction in an end-to-end manner.

\subsection{Planning}
\label{sec:related-planning}
As a critical component near the end of the autonomous driving pipeline, planning directly dictates the system's final performance.
Early algorithms employ various kinds of approaches including trajectory optimization \cite{diachuk2022motion, shi2022path}, graph search \cite{kim2014road, reda2024path}, sampling-based methods \cite{ma2015efficient, li2020adaptive, orthey2023sampling} to achieve specific objectives for planning. In the past few years, learning-based methods relying on reinforcement learning \cite{kendall2019learning, aradi2020survey, peng2022drl} have gradually emerged. Nevertheless, due to the significant gap between simulation and real-world environments, these methods often underperform in actual driving scenarios. Recently, some impressive works \cite{hu2023planning, jiang2023vad} attempt to realize planning for ego-vehicle in end-to-end manners. Impressively, UniAD\cite{hu2023planning} introduces a unified planning-oriented framework that integrates full-stack driving tasks in the dense BEV-Centric paradigm. VAD \cite{jiang2023vad} models the driving scene as a fully vectorized representation and exploits explicit instance-level planning constraints to improve planning safety.

\section{Method}
\label{sec:method}
\begin{figure}[tb]
  \centering
  \includegraphics[width=1\linewidth]{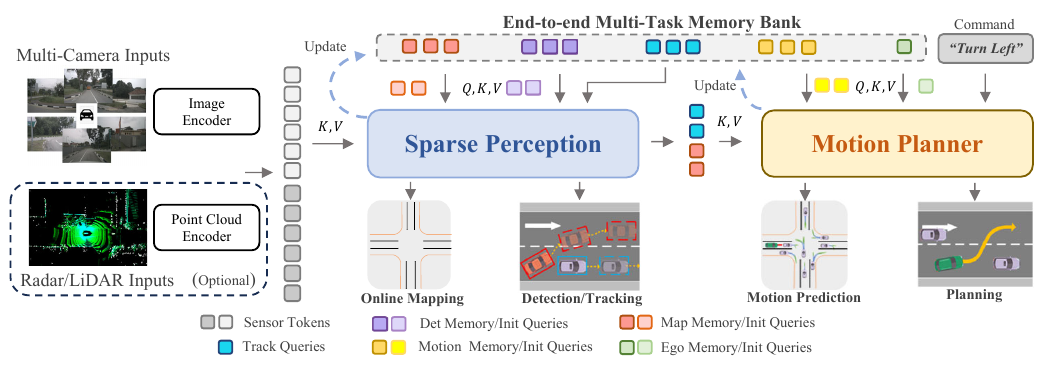}
  \caption{The pipeline of \textbf{\model}, in which sparse queries completely represent the driving scenario across space, time, and tasks without any dense BEV feature. }
  \label{fig:arch}
\end{figure}
\subsection{Overview}

As shown in \cref{fig:pipeline-c}, in the proposed sparse query-centric paradigm, different sparse queries completely represent the whole driving scenario, responsible for not only the information transfer and interaction across modules but also propagating backward gradients for optimization in multi-tasks in an end-to-end manner. Different from previous dense BEV-centric methods \cite{hu2023planning, jiang2023vad, ye2023fusionad}, there are not any view-projectors \cite{philion2020lift} and dense BEV features, which avoids the heavy computational and memory burden \cite{liu2022petr, jiang2023far3d, zhang2023fully}.

The detailed architecture of \model is shown in \cref{fig:arch}. Schematically, \model consists of three parts, including sensor encoders, sparse perception and motion planner. More specifically, sensor encoders take multi-view camera images, radar or LiDAR points as input and encode them into high-dimensional features, which are then fed into sparse perception module as sensor tokens with positional embedding (PE) \cite{liu2022petr, yan2023cross}. In sparse perception module, the raw data from sensors will be aggregated into multiple kinds of sparse perception queries such as detect queries, track queries and map queries which represent different elements of the driving scenario and will further be propagated to downstream tasks. In motion planner, perception queries are regarded as the sparse representation of driving scenarios and fully exploited for all surrounding agents and ego-vehicle, then driving constraints of multiple aspects are considered to generate the final planning which meets both safety and kinematic requirements. Moreover, the end-to-end multi-task memory bank is introduced in the architecture to uniformly store temporal information of the entire driving scenario, which makes it feasible to benefit from long-temporal historical aggregation for full-stack driving tasks. More details about the definition and implementation of the proposed method are provided in the \cref{appendix-imple}.

\subsection{Sparse Perception}
\label{sec:sparse_perception}
\begin{figure}[tb]
  \centering
  \includegraphics[width=1\linewidth]{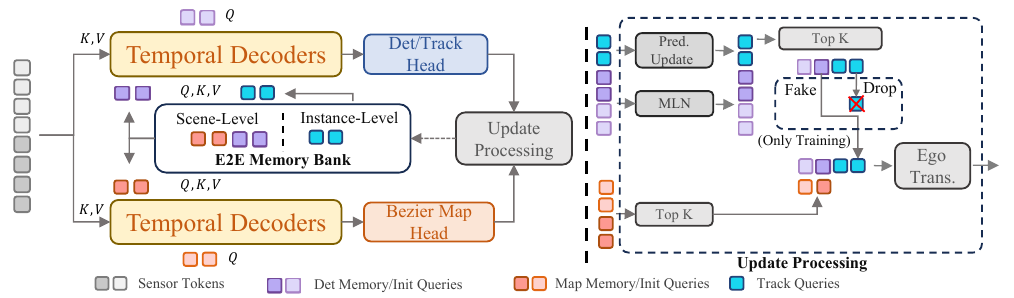}
  \caption{The module of \textbf{Sparse Perception}, which consists of two unified decoders and two corresponding heads for detection\&tracking and online mapping respectively. }
  \label{fig:dt-map-arch}
\end{figure}

As shown in \cref{fig:dt-map-arch}, the sparse perception module of \model unifies multiple perception tasks including detection, tracking and online mapping in the sparse manner. Concretely, there are two temporal decoders \cite{wang2023exploring} which share exactly the same structure to take advantage of long-term historical information from memory bank, where one is for obstacle perception and another for online mapping. Following the process of information aggregation via different perception queries of corresponding tasks, detection\&track head and Bezier map head are applied to decode and output obstacles and map elements respectively. After that, the update process which filters and saves high-confidence perception queries of current frame is carried out and the memory bank will be updated correspondingly, which will benefit the perception procedure in the next incoming frame.

\paragraph{\textbf{Detection\&Tracking.}} With regard to perception of obstacles, we adopt joint detection and tracking within the unified decoders without any additional hand-craft post-processing. As studied in \cite{zhang2023motrv2, yu2023motrv3}, there is an apparent imbalance between detect and track queries which may result in a non-negligible degradation of detection performance.

In order to alleviate the above issues, we improve the performance of obstacle perception from various perspectives. Firstly, we introduce a two-level memory mechanism to propagate temporal information across frames, in which scene-level memory maintains information from queries without any cross-frame associations while the instance-level memory keeps the corresponding relationship between adjacent frames of tracked obstacles. Secondly, we apply different update strategies for scene-level and instance-level memories considering the difference of their origins and missions. Specifically, we update scene-level memory via MLN \cite{wang2023exploring} while the instance-level memory is updated through the future predictions of each obstacle. Additionally, augmentation strategies \cite{zeng2022motr} on track queries are employed to balance the supervision between two-level memories during training so as to enhance detection and tracking performance.

Afterward, 3D bounding boxes with attributes and unique IDs can be decoded from detect or track queries through the detection\&track head and then can be further used in downstream tasks.

\paragraph{\textbf{Online Mapping.}} To the best of our knowledge, all of the existing methods of online mapping rely on dense BEV features to represent the driving environment, making it arduous to extend the perception scope larger or benefit from historical information for the high memory and computational cost. Based on the cognition and faith that all map elements can be represented in the sparse manner, we attempt to accomplish online mapping in the sparse paradigm.

Specifically, we adopt the temporal decoders which share the same structure as the ones equipped in obstacle perception tasks. Initially, map queries with priori classes are initialized to be uniformly scattered on the driving plane. Within the temporal decoders, map queries interact with the sensor tokens and historical memory tokens which actually consist of map queries with high confidence from previous frames. Then the updated map queries which carry valid information about map elements of current frame can be pushed into the memory bank so as to be used in future frames or downstream tasks. Obviously, the pipeline of online mapping is approximately the same as obstacle perception. That is to say, we unify perception tasks including detection, tracking, and online mapping in a general sparse manner which is much more efficient to extend to a larger scope (e.g., $100m\times100m$) or long-term fusion without any complex operators (e.g., deformable attention\cite{zhu2020deformable}, multi-point attention\cite{yuan2024streammapnet}). As far as we know, we are the first to implement online mapping within a unified perception architecture in the sparse manner.

Subsequently, we leverage the piece-wise Bezier \cite{qiao2023end} map head to regress the piece-wise Bezier control points of each sparse map element, which can be transformed to meet the requirements for downstream tasks conveniently.

\subsection{Motion Planner}
\label{sec:motion_planner}

In this paper, we revisit motion prediction and planning in the autonomous driving system and realize that many previous methods\cite{shi2024mtr++, zhou2022hivt} ignore ego-vehicle when making motion predictions for surrounding agents. Although in most cases this problem does not manifest itself, there may be some potential risks when driving in scenarios with strong interactions between near agents and ego-vehicle such as intersections.
Inspired by the above, we devise a more justifiable motion planner framework, in which motion predictions for surrounding agents and ego-vehicle are conducted simultaneously by the motion predictor. The predictions for ego-vehicle are then used as motion prioris in the subsequent planning optimizer.
Then constraints of different aspects are considered to produce the final planning results that satisfy both safety and kinematic requirements.

\begin{figure}[tb]
  \centering
  \includegraphics[width=1\linewidth]{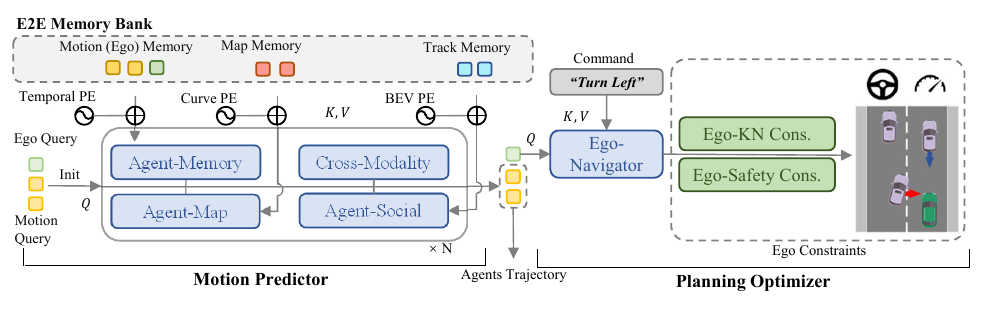}
  \caption{The module of \textbf{Motion Planner} considers the relationships among different agents and ego-vehicle, as well as driving constraints for safety and kinematics.}
  \label{fig:fore-arch}
\end{figure}

As shown in \cref{fig:fore-arch}, the motion planner in \model takes perception queries including track queries and map queries as the sparse representation of current driving scenario and the multi-modal motion queries are utilized as the medium to realize the awareness of driving scenario, the interaction between all agents including ego-vehicle, and the gaming among different future possibilities. Then the multi-modal motion queries of ego-vehicle would be sent into planning optimizer, where driving constraints of multiple aspects including high-level command, safety, and kinematics are fully considered in different ways.

\paragraph{\textbf{Motion Predictor.}} Following previous methods \cite{hu2023planning, ye2023fusionad}, the awareness and integration between motion queries and current driving scenario representation including track queries and map queries are realized through the vanilla transformer \cite{vaswani2017attention} layers. Besides, we apply ego-agent and cross-modality interaction to jointly model the interactions between surrounding agents and ego-vehicle in the future spatial-temporal scene. With the synergy of above modules within and between multi-layer stacked structures, motion queries are able to aggregate rich semantic information from both static and dynamic environment.

Apart from mentioned above, we introduce two strategies to further improve the performance of motion predictor. Firstly, we make a simple and straightforward prediction with the instance-level temporal memories of track queries and take it as part of the initialization for motion queries of surrounding agents. In this way, motion predictor is able to profit from the prioris obtained in upstream tasks. Secondly, benefit from the end-to-end memory bank, we assimilate useful information from the saved historical motion queries with the agent-memory aggregator with nearly negligible cost in a streaming manner.

It should be noted that the multi-modal motion queries for ego-vehicle are updated simultaneously. In this 
way it's possible to obtain the motion priori of ego-vehicle, which can further facilitate the learning process
of planning.

\paragraph{\textbf{Planning Optimizer.}} By means of the motion priori from motion predictor, we get a better initialization to take less detours during the training process. As a key component of our motion planner, the design of cost functions is crucial for it will greatly affect or even determine the quality of the final performance. In motion planner of the proposed \model, we consider different constraints from two major aspects, safety and kinematics, aiming to generate satisfying planning results.

Specifically, beyond the constraints identified in VAD\cite{jiang2023vad}, we focus on the dynamic safety relationships between ego-vehicle and nearby agents, considering their relative positions in the future.  For example, if agent $i$ persistently remains in the front-left area relative to ego-vehicle, thereby preventing leftward maneuvers of ego, agent $i$ will gain a \textit{left} label, indicating a leftward constraint imposed by agent $i$ on ego-vehicle (details in \cref{appendix-imple-motionplanner}). Accordingly, constraints are classified as \textit{front}, \textit{back}, or \textit{no} in the longitudinal direction, and as \textit{left}, \textit{right}, or \textit{no} in the lateral direction.
In motion planner, we decode the relationships between other agents and the ego vehicle in both lateral and longitudinal directions from the corresponding queries. This process involves determining the probabilities of all constraint relationships between other agents and the ego vehicle in these directions.
Then we utilize focal loss\cite{lin2017focal} as the cost function for the Ego-Agent Relationship (EAR), effectively capturing the potential risks from nearby agents:

\begin{equation}
    \mathcal{L}_{EAR} = - \sum_{d=1}^D\sum_{i=1}^A\sum_{c=1}^C\alpha_i(1-\hat{R^d_{ic}})^\gamma \log(\hat{R^d_{ic}})
\end{equation}
where $A$ represents the number of surrounding agents, $C$ denotes the number of classes, $D$ is the number of constraint directions (longitudinal and lateral) and $\alpha_i$, $\gamma$ are hyper-parameters in focal-loss.
We use \(\hat{R}^d_{ic}\) to denote the probability of the relationship \(c\) in direction \(d\) of agent \(i\).

Since the planning trajectories must comply with kinematic laws for control system execution, we embed auxiliary tasks in the motion planner to prompt the learning of the kinematic status of ego-vehicle. We decode status like velocity, acceleration, and yaw from the ego query $Q_{\text{ego}}$ and supervise these with a kinematic loss:

\begin{equation}
    L_{\text{KN}} = \frac{1}{N} \sum_{i=1}^{N} \left(\textrm{Dec}(Q_{\text{ego}})_i - \text{status}_i\right)^2
\end{equation}
where $N$ denotes the number of ego status, and $\text{status}_i$ denotes the $i$-th actual status of ego-vehicle.

\section{Experiments}
\label{sec:experiments}

We conduct substantial experiments on the challenging nuScenes \cite{caesar2020nuscenes} dataset to demonstrate the validity and superiority of our method. Impartially, the performance of each of the full-stack tasks will be evaluated to compare with previous methods whether in an end-to-end manner or not. In summary, the experiments in this section use three different configurations of SparseAD, which are SparseAD-B and SparseAD-L, which use only image inputs, and SparseAD-BR, which uses radar point cloud and image multi-modal inputs. Both SparseAD-B and SparseAD-BR utilize the V2-99\cite{lee2020centermask}  as the image backbone network with input images in a resolution of $1600\times640$. SparseAD-L further leverages ViT-Large\cite{dosovitskiy2020image} as the image backbone with input images in a resolution of $1600\times800$. More details about the experiments are provided in the \cref{appendix-exps}.

\subsection{Comprehensive Results}
\begin{table*}[t]
\caption{A comprehensive comparison of SparseAD with the state-of-the-art full-task end-to-end autonomous driving method on multi-task performance. Ped.: Pedestrian crossing, Div: Divider, T.Mem: Training memory footprint.}
\begin{center}
\tiny
\vspace{-10pt}
\resizebox{1\textwidth}{!}
{
\begin{tabular}{l|c|cc|cc|cc|ccc|cc|ccc}
\toprule
\multirow{2}{*}{Method} &
\multirow{2}{*}{Modal} &
\multicolumn{2}{c|}{Detection} & 
\multicolumn{2}{c|}{Tracking} & 
\multicolumn{2}{c|}{Online Mapping} & 
\multicolumn{3}{c|}{Motion Prediction} & 
\multicolumn{2}{c}{Planning } &
\multicolumn{3}{c}{Efficiency} \\
&& mAP$\uparrow$ & NDS$\uparrow$ & AMOTA$\uparrow$&Recall$\uparrow$ &IoU-Ped.$\uparrow$ & Iou-Div.$\uparrow$& minADE$\downarrow$&minFDE$\downarrow$&MR$\downarrow$ &avg.L2$\downarrow$ &avg.Col.$\downarrow$ & Params & T. Mem & FPS\\
\midrule
UniAD\cite{hu2023planning}&C&0.380& 0.499& 0.359& 0.467&0.138&0.257&1.17& 1.65& 0.205& 1.03 &0.31&124&50.0&1.8\\
\midrule
SparseAD-B&C& 0.475& 0.578& 0.530& 0.608 &0.164&0.288& 0.83& 1.58& 0.187& 0.35& 0.09&124&17.5&3.5\\
SparseAD-BR&C+R&0.480& 0.579 & 0.534 &0.635&0.165&\textbf{0.289}&0.81& 1.55 &0.183 &\textbf{0.34}& 0.08&125&17.6&3.5\\
SparseAD-L&C&\textbf{0.536}& \textbf{0.625} & \textbf{0.606}& \textbf{0.706}&\textbf{0.168}&0.278&\textbf{0.74}& \textbf{1.46}& \textbf{0.169}& \textbf{0.34}& \textbf{0.06}&376&31.6&0.7\\
\bottomrule
\end{tabular}% }
}
\end{center}
\label{tab:joint-results}
\end{table*}

As shown in \cref{tab:joint-results}, \model achieves much more excellent performance in all tasks with the speed of nearly $\mathbf{1 \boldsymbol{\times}}$ faster, and \textbf{35\%} memory footprint compared with UniAD \cite{hu2023planning}, which possesses the capability of full-stack driving tasks and is far-reaching in the field of end-to-end autonomous driving. 

Concretely, we improve the metrics of mAP, NDS, AMOTA, Recall to \textbf{47.5\%}, \textbf{57.8\%}, \textbf{53\%}, \textbf{60.8\%} respectively with only multi-view camera inputs, which evidently proves the superiority of sparse architecture for obstacle perception once again. Also for online mapping, \model has also obtained better performance than UniAD \cite{hu2023planning} in which BEV features are utilized for dense representation of the driving scenario. As for the tasks of motion prediction and planning, \model also achieves significantly better results, with the minADE, average L2 error and Collision Rate substantially reduced to \textbf{0.83m}, \textbf{0.35m} and \textbf{0.09\%}, respectively. We believe that it comes from not only the superior performance of upstream perception tasks but also the rational design of motion planner.

Moreover, the performance of \model has been efficiently enhanced by incorporating multi-modal sensor inputs from multi-view cameras and radar, with almost negligible extra training and computation costs. Besides, with the advanced backbone network of ViT-Large\cite{dosovitskiy2020image}, the comprehensive performance further improved significantly, which strongly proved the potential of the sparse query-centric paradigm by scaling up to achieve better performance.

\subsection{Multi-Task results}
\begin{table}[t]
\centering
\caption{Results of 3D detection and 3D multi-object tracking on the nuScenes \texttt{val} dataset. "Tracking-only methods" stands for the methods tracking by post-processing association. "End-to-End Autonomous Driving Method" stands for the method with autonomous driving full-stack task capability. All methods in the table are evaluated under full-resolution image inputs. $^\dag$: The results are reproduced by the official open-source code. -R: with radar point cloud inputs.}
\label{tab:det_tracking_val_set}
\tiny
\resizebox{\textwidth}{!}{
% {
\setlength{\tabcolsep}{3.5pt}
\begin{tabular}{l|c|c|ccc|l|cccc} 

\toprule
\textbf{Methods}&  \textbf{mAP}$\uparrow$  &\textbf{NDS}$\uparrow$  & \textbf{mATE}$\downarrow$  &\textbf{mAOE}$\downarrow$   &\textbf{mAVE}$\downarrow$ & \textbf{Methods}& \textbf{AMOTA}$\uparrow$  & \textbf{AMOTP}$\downarrow$ & \textbf{RECALL}$\uparrow$ & \textbf{IDS}$\downarrow$\\
\midrule 
\multicolumn{5}{l}{\textit{Detection-only methods}}&&\multicolumn{5}{l}{\textit{Tracking-only methods}}\\
\midrule
% ResNet101
StreamPETR\cite{wang2023exploring}                            &0.504  &0.592  &0.569    &0.315  &0.257  &DEFT\cite{chaabane2021deft} & 0.201 & - & - & -\\ 
SOLOFusion\cite{park2022time}                & 0.483 & 0.582 & \textbf{0.503}  & 0.381 & 0.246 &QD3DT\cite{hu2022monocular} & 0.242 & 1.518 & 0.399  & 5646 \\ 
BEVDepth\cite{li2023bevdepth}                         & 0.412 & 0.535 & 0.565  & 0.358 & 0.331 &TripletTrack~\cite{marinello2022triplettrack} & 0.285 & 1.485 &-  & - \\ % PETR Table 1
BEVFormer \cite{li2022bevformer}                  & 0.416 & 0.517 & 0.673  & 0.372 & 0.394 & CC-3DT$^*$~\cite{fischer2022cc} & 0.429 & 1.257 & 0.534 & 2219 \\
Sparse4D\cite{lin2022sparse4d}                 & 0.436 & 0.541 & 0.633  & 0.363 & 0.317  &QTrack\cite{yang2022quality} &0.511& \underline{1.090}& 0.585 & 1144\\ 
SRCN3D\cite{shi2022srcn3d}& 0.396& 0.475& 0.737& \underline{0.278}& 0.728& SRCN3D\cite{shi2022srcn3d} &0.439 &1.280& 0.545&-
\\ 

\midrule
\multicolumn{7}{l}{\textit{End-to-End Multi-Object Tracking Methods}}\\
\midrule
MUTR3D\cite{zhang2022mutr3d}  &-&-&-&-&-& MUTR3D\cite{zhang2022mutr3d} &0.294& 1.498& 0.427 &  3822 \\
PF-Track$^\dag$\cite{pang2023standing} &	0.382& 0.486&	0.702&	0.430&	0.463 &PF-Track$^\dag$\cite{pang2023standing} & {0.479} & {1.227} & {0.590}  & \textbf{181}\\
Sparse4Dv3\cite{lin2023sparse4d} &\textbf{0.537} &\underline{0.623}&\underline{0.511} & 0.306 &\textbf{0.194} & Sparse4Dv3\cite{lin2023sparse4d} &\underline{0.567}& 1.027 & \underline{0.658}& 557\\
\midrule
\multicolumn{7}{l}{\textit{End-to-End Autonomous Driving Methods}}\\
\midrule
UniAD\cite{hu2023planning} &0.380&0.499&0.684	&	0.383	&0.378& UniAD\cite{hu2023planning} &0.359 &1.320 &0.467 &906\\ 
\rowcolor[gray]{.9}  \model-B& 0.475 & 0.578&0.556	&	0.295&	0.288	& \model-B & 0.530	& 1.087&	0.608 & 297\\
\rowcolor[gray]{.9}  \model-BR& 0.480 & 0.579 &0.554	&	0.316&	0.273	& \model-BR & 0.534	& 1.077&	0.635 & 277\\
\rowcolor[gray]{.9}  \model-L& \underline{0.536} & \textbf{0.625} &0.531	&	\textbf{0.219}&	\underline{0.230}	& \model-L & \textbf{0.606}	& \textbf{0.933}&	\textbf{0.706} & \underline{251}\\
\bottomrule
\end{tabular}}
\vspace{-0.25cm}
\end{table}

\paragraph{\textbf{Obstacle Perception.}}
We compare the detection and tracking performance of \model with other methods on the nuScenes \texttt{val} set in \cref{tab:det_tracking_val_set}. Obviously, \model-B outperforms most of the popular detection-only, tracking-only and end-to-end multi-object tracking methods \cite{li2023bevdepth, li2022bevformer, pang2023standing,zhang2022mutr3d}, while maintaining a comparable performance on the corresponding task when compared to the SOTA methods such as StreamPETR\cite{wang2023exploring}, QTrack\cite{yang2022quality}. With the scale-up via advanced backbone, \model-Large achieves overall better performance with \textbf{53.6\%} mAP, \textbf{62.5\%} NDS, and \textbf{60.6\%} AMOTA, which is overall better than Sparse4Dv3 \cite{lin2023sparse4d}, the previous state-of-the-art.

\begin{table}[t]
	\caption{Performance comparison with the online mapping method. The results are evaluated under thresholds with $[1.0m, 1.5m, 2.0m]$. $\ddag$: Reproduced by the official open-source code. $\dag$: Based on the requirement of the planning module in \model, we further divide \textit{boundary} into \textit{Road Segment} and \textit{Lane} which are evaluated separately. $*$: The cost of the backbones and sparse perception module. -R: with radar point cloud inputs.}
        \vspace{-10pt}
        % \scriptsize
        \tiny
	\begin{center}
		% \resizebox{1.0\textwidth}{!}
            {
			\begin{tabular}{l|p{1.3cm}<{\centering}p{1.3cm}<{\centering}p{2cm}<{\centering}p{1cm}<{\centering}p{3.0cm}<{\centering}}
				\toprule
				Method & \textbf{Divider}$\uparrow$& \textbf{Cross}$\uparrow$ & \textbf{Boundary}$\uparrow$ & mAP$\uparrow$ & Train Memory(GB)$\downarrow$\\
                \midrule 
                \multicolumn{6}{l}{\textit{Online-Mapping-Construction-only methods}}\\
                \midrule
				HDMapNet$^\ddag$ \cite{li2022hdmapnet} & 18.0 & 23.1 & 33.5 & 24.9  & 21.1\\
				VectorMapNet$^\ddag$ \cite{liu2023vectormapnet}  & 39.8 &  38.7 & 19.9 & 32.8 &16.4\\
				MapTR$^\ddag$ \cite{liao2022maptr}  & 17.9 &  36.9 & 1.8 & 18.9 & 20.0\\
			  StreamMapNet$^\ddag$ \cite{yuan2024streammapnet} & 55.5 &  59.7 & 46.7 & 54.0 &23.2\\
				\midrule 
                \multicolumn{6}{l}{\textit{End-to-End Multi-Object Tracking Methods}}\\
                \midrule
				\rowcolor[gray]{.9} {SparseAD-B}  & {37.3}& {45.2} & {29.2/25}$^\dag$ & {34.2} & \textbf{6.9}$^*$\\
                    \rowcolor[gray]{.9} {SparseAD-BR}  & {38.4}& {44.5} & {29.4/24.7}$^\dag$ & {34.2} & 7.1$^*$\\
                    \rowcolor[gray]{.9} {SparseAD-L}  & {37.6}& {46.7} & {30/24.8}$^\dag$ & {34.8} & 19.3$^*$\\
                \bottomrule
			\end{tabular}
		}
	\end{center}
	\label{tab:map-nuscenes-val}
         \vspace{-10pt}
\end{table}

\paragraph{\textbf{Online Mapping.}} We display the comparison result of online mapping between \model and other previous methods in \cref{tab:map-nuscenes-val} on the nuScenes \texttt{val} set. It should be pointed out that we divide the \textit{boundary} into \textit{road segment} and \textit{lane} which are evaluated separately and extend the range from typically $60m \times 30m$ to $102.4m \times 102.4m$ to be consistent with obstacle perception, according to the needs of planning. Without losing fairness, \model achieves \textbf{34.2\%} mAP in the sparse end-to-end manner without any dense BEV representations, which outperforms most of previous popular methods such as HDMapNet \cite{li2022hdmapnet}, VectorMapNet \cite{liu2023vectormapnet} and MapTR \cite{liao2022maptr} with obvious advantages in terms of both performance and training costs. Although the performance lags behind StreamMapNet \cite{yuan2024streammapnet}, it is demonstrated that online mapping could be accomplished in a unified sparse manner without any dense BEV representation, which is of great significance for the real-world deployment of end-to-end autonomous driving with significantly lower cost. Admittedly, how to effectively utilize the useful information from other modalities such as Radar is still a task that deserve further exploration. We believe there is still lots of room to be further explored in the sparse manner.

\paragraph{\textbf{Motion prediction.}} We display the comparison result of motion prediction in \cref{tab:motion_vip3d_val_set} where the metrics are consistent with VIP3D \cite{gu2023vip3d}. \model achieves the best performance with lowest \textbf{0.83m} minADE, \textbf{1.58m} minFDE, \textbf{18.7\%} Miss Rate and highest \textbf{0.308} EPA among all end-to-end methods \cite{liang2020pnpnet, gu2023vip3d, hu2023planning} with huge advantage. Furthermore, thanks to the efficiency and scalability of the sparse query-centric paradigm, \model could be expanded to more modalities efficiently and benefit from advanced backbones, which further improves the prediction performance markedly.

\begin{table}[t]

    \centering
    % \scriptsize
    \caption{The comprehensive performance of motion planner in SparseAD on the nuScenes \texttt{val} dataset. -R: with radar point cloud inputs.}
    \vspace{-5pt}
    \begin{subtable}[h]{0.45\textwidth}
    \tiny
         
    {
    \begin{tabular}{l|cccc}
		\toprule
		Method & minADE$\downarrow$ & minFDE$\downarrow$ & MR$\downarrow$ & EPA$\uparrow$ \\
		\midrule
            Traditional\cite{gu2023vip3d} &2.06&3.02&0.277&0.209\\
		PnPNet~\cite{liang2020pnpnet} & 2.04 & 3.03 & 0.271 & 0.213 \\
		ViP3D~\cite{gu2023vip3d} & 2.03 & 2.90 & 0.239 & 0.236 \\
		{UniAD}\dag~\cite{hu2023planning} & 1.17&	1.65&	0.205&	0.205 \\
        \midrule
        \rowcolor[gray]{.9} \model-B & 0.83&	1.58	&0.187& 0.308\\
        \rowcolor[gray]{.9} \model-BR & 0.81&	1.55&	0.183& \textbf{0.316}\\
         \rowcolor[gray]{.9} \model-L &\textbf{0.74}	&\textbf{1.46}&	\textbf{0.169}& 0.292\\
		\bottomrule
	\end{tabular}
        }
        \caption{Comparing motion prediction performance with previous vision-based end-to-end methods on ViP3D\cite{gu2023vip3d} benchmark. $\dag$: Evaluated from official checkpoint\cite{hu2023planning}.}
	\label{tab:motion_vip3d_val_set}
    \end{subtable}
    \hfill
    \begin{subtable}[h]{0.50\textwidth}
    \tiny
    
    \begin{tabular}{l|cccc|cccc}
    \toprule
    \multirow{2}{*}{Method} &
    \multicolumn{4}{c|}{L2 (m) $\downarrow$} & 
    \multicolumn{4}{c|}{Collision (\%) $\downarrow$}  \\
    
    & 1s & 2s & 3s & Avg. & 1s & 2s & 3s & Avg. \\
    \midrule
    EO$^\dagger$~\cite{khurana2022differentiable}& 0.67 & 1.36 & 2.78 & 1.60 & 0.04 & 0.09 & 0.88 & 0.33   \\
    \midrule
    ST-P3~\cite{hu2022st}& 1.33 & 2.11 & 2.90 & 2.11 & 0.23 & 0.62 & 1.27 & 0.71  \\
    UniAD~\cite{hu2023planning} & 0.48 & 0.96 & 1.65 & 1.03 & 0.05 & 0.17 & 0.71 & 0.31  \\
    VAD~\cite{jiang2023vad} & 0.17 & 0.34 & 0.60 & 0.37 & 0.07 & 0.10 & 0.24 & 0.14  \\
    \midrule
    \rowcolor[gray]{.9}  \model-B&\textbf{0.15}&\textbf{0.31}&0.57&0.35 &	\textbf{0.00}&0.06&0.21&0.09\\
    \rowcolor[gray]{.9}  \model-BR&0\textbf{.15}&\textbf{0.31}&\textbf{0.56}&\textbf{0.34}&	\textbf{0.00}&0.06&0.19&0.08\\
    \rowcolor[gray]{.9}  \model-L&\textbf{0.15}&\textbf{0.31}&\textbf{0.56}&\textbf{0.34}&	\textbf{0.00}&\textbf{0.04}&\textbf{0.15}&\textbf{0.06}\\
    \bottomrule
    \end{tabular}
    \caption{Open-loop planning performance on the nuScenes \texttt{val} dataset. LiDAR-based methods are denoted with \dag.}
    \label{tab:planning_val_set}
    \end{subtable}
    \vspace{-20pt}
\end{table}

\paragraph{\textbf{Planning.}} The results of planning are presented in \cref{tab:planning_val_set}. Thanks to the superior design of both the upstream perception modules and the motion planner, all versions of \model have achieved state-of-the-art on the nuScenes \texttt{val} dataset. Specifically, \model-B achieves the lowest average L2 error and Collision Rate compared to all other methods including UniAD\cite{hu2023planning} and VAD\cite{jiang2023vad} which demonstrate the superiority of our approach and architecture. Similar to upstream tasks including obstacle perception and motion prediction, \model gains further improvement with radar or more powerful backbone networks.

\subsection{Ablations}

\begin{table*}[t]
\caption{Ablation for designs in the sparse perception module. "Det. Mem." means Detection Scene-Level Memory; "Map. Mem." means Map Scene-Level Memory; "Two-L. M. U." means Two-Level Memory Update; "TQeury Aug." means Track Query Augmentation;
}
\begin{center}
\tiny
\vspace{-10pt}
\resizebox{0.9\textwidth}{!}
{
\begin{tabular}{l|cccc|cc|ccc|c}
\toprule
\multirow{2}{*}{ID} &
\multirow{2}{*}{\makecell{Det. \\ Mem.}} &
\multirow{2}{*}{\makecell{Map. \\ Mem.}} &
\multirow{2}{*}{\makecell{Two-L.\\ M. U.}} &
\multirow{2}{*}{\makecell{TQeury\\ Aug.}} &
\multicolumn{2}{c|}{Detection} & 
\multicolumn{3}{c|}{Tracking} &
{Mapping} \\

&&&& & mAP & NDS & AMOTA & AMOTP & Recall  & mAP  \\
\midrule
1& & & & & 0.4227	&0.5326	&0.443	&1.301&	0.538	&0.311\\
2&\Checkmark & & & & 0.4261&	0.5451&	0.42&	1.301	&0.525	&0.317\\
3&\Checkmark &\Checkmark & & & 0.4201&0.5414&0.415&1.316&0.513& 0.331\\
4&\Checkmark &\Checkmark &\Checkmark & & 0.4294&	0.5457&	0.439&	1.272&	0.535	&	\textbf{0.337}\\
5&\Checkmark &\Checkmark &\Checkmark &\Checkmark  &\textbf{0.4537}&	\textbf{0.5634}&	\textbf{0.472}&	\textbf{1.209}&	\textbf{0.58}	&	0.334\\ 
\bottomrule
\end{tabular}% }
}
\end{center}
\label{tab:perception-ablations}
\end{table*}

\paragraph{\textbf{Effect of designs in Perception.}} We provide ablation experiments for the module of sparse perception in \cref{tab:perception-ablations}. The experimental results show that introducing scene-level memory for detection and mapping can effectively improve the performance of corresponding tasks. However, the tracking performance is slightly degraded since the supervision of different queries may be unstable during training. With the help of update strategies of the two-level memories and augmentation strategies on track queries, the supervision of different queries gets to be more balanced and the performance of all perception tasks has been further improved.

\label{abalation:perceptrion}

\begin{table*}[t]
\caption{Ablation for designs in the motion planning module. "MQI" means Motion Query Init; "AMI" means Agent-Memory Interaction; "KN" means Ego Kinematic Constraint; "EAR" means Ego-Agent Relationship Constraint.}
\begin{center}
\tiny
\vspace{-10pt}
\resizebox{0.9\textwidth}{!}
{
\begin{tabular}{l|cccc|ccc|cccc|cccc}
\toprule
\multirow{2}{*}{ID} &
\multirow{2}{*}{\makecell{MQI}} &
\multirow{2}{*}{\makecell{AMI}} &
\multirow{2}{*}{\makecell{KN}} &
\multirow{2}{*}{\makecell{EAR}} &

\multicolumn{3}{c|}{Prediction} & 
\multicolumn{4}{c|}{Planning L2(m)} &
\multicolumn{4}{c|}{Planning Coll.(\%)} \\

&&&& & minADE&	minFDE	&MR & 1s & 2s  & 3s&avg. & 1s & 2s  & 3s&avg.\\
\midrule
1&\Checkmark&\Checkmark&\Checkmark&\Checkmark&0.876&	1.654&	0.193&	\textbf{0.35}&\textbf{0.69}&\textbf{\textbf{1.13}}&\textbf{0.72}&0.02& \textbf{0.22} & \textbf{\textbf{0.66}}&\textbf{0.30}\\
2&&\Checkmark&\Checkmark&\Checkmark&0.867&	1.660&	0.203	&\textbf{0.35} & 0.70 & 1.14	& 0.73 & 0.05 & 0.29&0.75 & 0.36\\
3&\Checkmark&&\Checkmark&\Checkmark& 0.870&1.646&	0.193	&0.39&0.76&1.23&0.79&	0.03&0.45&1.27&0.58\\
4&\Checkmark&\Checkmark&&\Checkmark&0.866&	1.646&	0.197	&0.36&0.71&1.15&0.74&	\textbf{\textbf{0.01}}&0.28&0.84&0.38\\
5&\Checkmark&\Checkmark&\Checkmark&&0.864&	1.646&0.197&	\textbf{0.35}&\textbf{0.69}&\textbf{1.13}&\textbf{0.72}	&0.03&0.23&0.70&0.32\\
\bottomrule
\end{tabular}% }
}
\end{center}
\label{tab:motionpalnner-ablations}
\end{table*}

\paragraph{\textbf{Effect of designs in Motion Planning.}} What needs to be pointed out is that we excluded the ego status from input to prevent shortcut learning in open-loop planning scenarios \cite{li2023ego}. As shown in \cref{tab:motionpalnner-ablations}, the performance of planning would decline to varying degrees without any strategies in motion predictor or the introduced constraints in planning optimizer, which strongly proves the correctness and effectiveness of our method.

\label{abalation:planning}

\subsection{Qualitative Results}
\begin{figure}[tb]
  \centering
  \includegraphics[width=1\linewidth]{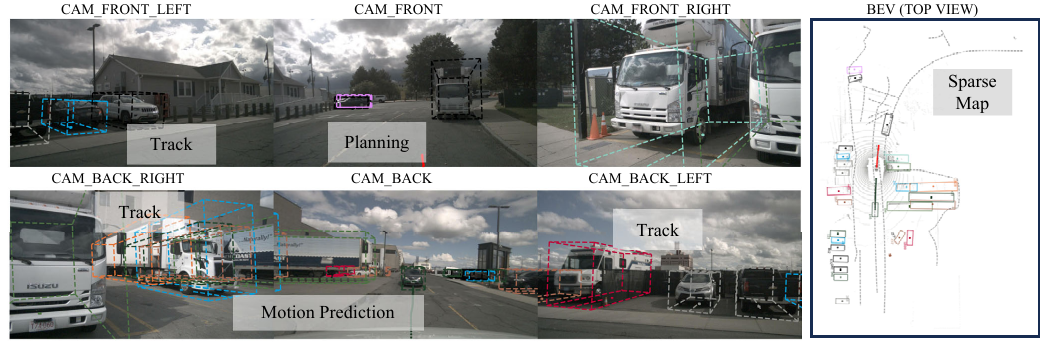}
  \caption{Qualitative visualization. We plot the results for all tasks in surround-view images and BEV. Each agent is marked by a dashed bounding box with different colors in both images and BEV. The sparse predictions of map elements are shown by gray dashed lines only in the BEV. The
top-1 modal of motion prediction and ego-vehicle planning results are selected for visualization on images and BEV respectively. The LiDAR point cloud in the BEV is only used in visualization which is not taken as inputs of the model.}
  \label{fig:vis}
\end{figure}

Intuitively, we display the performance of all tasks via the visualization as shown in \cref{fig:vis}, in which the challenging driving scenario is characterized by the presence of agents close to both the front and rear of ego-vehicle, as well as amounts of obstacles with potential occlusions on roadsides, and irregular road boundaries. Even so, SparseAD still produces an accurate perception of the complex elements in the whole scenario and provides a reasonable and safe planning trajectory. More visualizations and analysis of other driving scenarios can be found in \cref{appendix-vis}.

\section{Discussion}
Collision rate is a key metric for evaluating the performance of end-to-end autonomous driving methods. However, we find some serious disadvantages of the existing calculation methods of collision rate. We discuss this issue in detail in the \cref{appendix-discussion} and propose a new calculation method that is more precise and sensible.
\section{Conclusion and Future Work}
\paragraph{\textbf{Conclusion.}}
In this paper, we propose a novel sparse query-centric paradigm for end-to-end autonomous driving named SparseAD, which condenses the spatial and temporal information of the whole driving scenario into sparse queries. We revisit the tasks of autonomous driving and devise a more succinct architecture consisting of sparse perception and motion planner. The substantial experiments conducted on the challenging nuScenes dataset demonstrate the effectiveness and superiority of SparseAD. We hope SparseAD can reveal the great potential of end-to-end autonomous driving methods in performance, efficiency, and scalability.

\paragraph{\textbf{Limitations and future work.}}

Although SparseAD significantly improves the performance of full-stack driving tasks, how to balance multiple tasks during the optimization process is still worth exploring. In addition, further boosting the performance of online mapping in the sparse manner is a potential research field in the future.

\bibliographystyle{splncs04}
\bibliography{main}

\newpage
\appendix

\noindent\textbf{\large{Appendix}}

\section{Overview}
\label{appendix-overview}
We first present the outline of this supplementary material. \cref{appendix-definition} contains the definition of each task in SparseAD. \cref{appendix-imple} describes the implementation details of SparseAD. In addition to the experiments in the main manuscript, We provide more supplementary experiments to further validate the proposed method in \cref{appendix-exps}.  We further propose the variant paradigm of SparseAD in \cref{appendix-variant} to explore the trade-off of the perception capability between the precision\&recall and the temporal consistency. Specifically, in \cref{appendix-discussion}, we discuss in detail the disadvantages of existing occupancy-based collision rate evaluation metrics for end-to-end autonomous driving methods and propose a sparse-based method to calculate collision rates more precisely. We hope that the proposed metrics can be helpful and inspiring for future research and evaluation of end-to-end autonomous driving methods. Finally, extensive qualitative visualizations on various cases are shown in \cref{appendix-vis}.
\section{Definition of Each Task}
\label{appendix-definition}

\subsubsection{Detection and Tracking.}

As integral components of autonomous driving systems, detection and tracking enable the ego-vehicle to locate and perceive obstacles in the surrounding environment. Specifically, the module of detection needs to perceive obstacles around the ego-vehicle and provides information such as the position, size and orientation relative to the ego-vehicle. Furthermore, in the module of tracking, the temporal extension of detection, each obstacle is assigned with an unique ID, which should remain consistent across frames.

\subsubsection{Online Mapping.}

Aiming to replace the offline annotated High-Definition map, online mapping is responsible for modeling the structure and semantic information of the driving environment, which could significantly save labor costs and expand the applicability range of the autonomous driving system. Specifically, we regard the map as a sparse representation of line segments, with each road element corresponding to a set of sparse line segments, with a specific category such as dividers, pedestrian crosswalks, road boundaries, and lane lines.

\subsubsection {Motion Prediction.}

The module of motion prediction is responsible for predicting the possible future states of all traffic participants, so as to assist the procedure of decision making of the ego-vehicle. Generally, historical information is necessary to predict future trajectories for surrounding agents. Moreover,  maps are also an integral component of the input to the motion prediction module. It should be noted that aggregating and extracting information from the surrounding environment is beneficial for not only motion prediction for surrounding agents but also planning of ego-vehicle.

\subsubsection{Planning.}

As the component near the end of the autonomous driving pipeline, planning aims to provide a driving trajectory which should meet safety and comfort requirements, while also complying with kinematic laws. Under the guidance of the above goals, the planning module takes all information, consisting of current and possible future states of the surrounding environment and agents, from multiple upstream tasks as input, and fully considers different kinds of constraints to generate a collision-free, comfortable, and controllable trajectory.

\section{Implementation details}
\label{appendix-imple}
\subsection{Problem Definition}
\label{appendix-imple-def}

\subsubsection{Overview.}
\label{appendix-imple-def-overview}

The proposed SparseAD is compatible with inputs from multi-modal sensors, including cameras, millimeter-wave radar, and LiDAR. In this paper, the camera images are taken as the main input modality while millimeter-wave point clouds can be further introduced for an extra performance boost. In general, we denote the input images $\textbf{I}_1,... ,\textbf{I}_{N_i}$ jointly as $\textbf{I} \in \mathcal{R}^{N_i \times H \times W \times 3}$, where $N_i$ represents the number of input images; and denote the input point cloud as $\mathbf{P} \in \mathcal{R}^{N_p \times {C_p}}$, where $N_p$ represents the number of points in the input point cloud and $C_p$ is the dimension of the raw point cloud, which is 4 for LiDAR and 18 for radar in the nuScenes dataset. As shown in \cref{fig:arch}, the images and point clouds are encoded to a high-dimensional feature space with a dimension of $C$ by different encoders $E_i$ and $E_p$. We unify the encoded multi-sensor features into Sensor Tokens $\textbf{F}_t \in \mathcal{R}^{(N_i \times H \times W + N_p) \times C}$, which can be expressed as follows:
\begin{equation}
{{\mathbf{F}}_t} = {\rm{Concat}}\left( {{\rm{Flatten(}}{{\mathbf{F}}_I}{\rm{),Flatten(}}{{\mathbf{F}}_P}{\rm{)}}} \right)
\end{equation}
where
$$
\mathbf{F}_I = E_i(\mathbf{I}),\qquad\mathbf{F}_p = E_p(\mathbf{P})
$$

Similarly, the position embedding(PE) of Sensor Tokens is also obtained by concatenating the PEs of the tokens of each modality. In SparseAD, we use 3D-PE and BEV-PE for image features and point cloud features respectively, and its details can be referred to PETR\cite{liu2022petr} and CMT\cite{yan2023cross}.

In SparseAD, we use a variety of queries to achieve sparse representation for the driving scenario and maintain the multi-task capability. As shown in \cref{fig:arch}, most kinds of the queries can be divided into two categories: the initialized query which has learnable weights to acquire a priori features of the target distribution in training, and the memory queries from the End-to-End Multi-Task Memory Bank (EMMB) which is used to propagate the temporal information of each task, as well as to convey the upstream information for downstream tasks. We use the superscripts $i$ and $m$ to denote the learnable query $Q^i$ and memory query $Q^m$ respectively, and we will introduce them in the following parts. Note that here we introduce a two-level memory mechanism, in which the storage and update strategies are different for scene-level and instance-level memories, which will be described in detail in \cref{appendix-imple-summary}.

\subsubsection{Sparse Perception.}
\label{appendix-imple-def-perception}

As shown in \cref{fig:dt-map-arch}, in the sparse perception module, we set learnable detection queries $\mathbf{Q}^i_d\in \mathcal{R}^{N^i_d \times C}$ and online mapping queries $\mathbf{Q}^i_o\in \mathcal{R}^{N^i_o \times C}$ for realizing the solution of basic perception of obstacles and map elements in the current driving scene, where $N^i_d$ and $N^i_o$ are the number of two kinds of query, respectively. Meanwhile, the temporal information from the EMMB is also introduced as memory queries, which consist of scene-level memory including detection memory queries $\mathbf{Q}^m_d\in \mathcal{R}^{N^m_d \times C}$ and mapping memory queries $\mathbf{Q}^m_o\in \mathcal{R}^{N^m_o \times C}$, as well as track query as instance-level memory $\mathbf{Q}^m_{t,T-1}\in \mathcal{R}^{N^m_i \times1\times C}$, where $N^m_d,N^m_o$ is the length of detection and online mapping memory respectively, $N^m_i$ is the number of memory instance in the EMMB, and $T-1$ denotes the last frame.

Finally, the sparse perception module outputs detailed information about both the obstacles (coordinates, size, ID, velocity, etc.) and map elements (Bezier control points, sampling points, etc.)  in the driving scene, as well as the corresponding context-rich queries, which is provided to the downstream Motion Planner and updated into the EMMB for the perception in future frames. The implementation details of the sparse perception module will be described in detail in \cref{appendix-imple-perception}.

\subsubsection{Motion Planner.}
\label{appendix-imple-def-motionplanner}

As shown in \cref{fig:fore-arch}, we unified the multi-modal motion prediction for both surrounding agents and ego-vehicle in Motion Planner. Since the number of motion queries variants with the number of instances, we set a shared learnable embedding $\mathbf{E}_n \in \mathcal{R}^{1\times C}$ for all motion queries. We simply encode the historical trajectories $\tau_h$  of each instance by an MLP followed by a cross-attention layer to aggregate the historical perceptual information of the instance (e.g., track query $\mathbf{Q}^m_{t}\in \mathcal{R}^{N^m_i \times M\times C}$, where $M$ is the max history length of instances) and combine it with the learnable embedding $\mathbf{E}_n$ as the initialization of motion query $\mathbf{Q}^i_n \in \mathcal{R}^{N^m_i \times C}$. In particular, we only encode the ego historical trajectory $\tau_{eh}$ and combine it with a special ego embedding $\mathbf{E}_e \in \mathcal{R}^{1\times C}$ as the initialization of the ego query $\mathbf{Q}^i_e \in \mathcal{R}^{1 \times C}$. Similarly, we introduce the memory motion query of instances $\mathbf{Q}^m_n \in \mathcal{R}^{N^m_i \times M\times C}$, ego $\mathbf{Q}^m_e \in \mathcal{R}^{1 \times M\times C}$, and the sparse representation of agents $\mathbf{Q}^m_{t,T}\in \mathcal{R}^{N^m_i \times1\times C}$ and map elements $\mathbf{Q}_{o}\in \mathcal{R}^{N^i_o \times C}$ of the current frame from sparse perception as tokens, and fully interact with motion queries and ego query to generate high-quality multi-modal motion prediction trajectories $\hat{\tau}$. Subsequently, the ego query is further fed into Planning Optimizer with full consideration of safety and kinematic constraints to finally generate safe and reliable ego-vehicle trajectory planning  $\hat{\tau_e}$. The implementation details of the motion planner will be described in detail in \cref{appendix-imple-motionplanner}.
\subsection{Sparse Perception}
\label{appendix-imple-perception}

As shown in \cref{fig:dt-map-arch}, the Sparse Perception module takes sensor tokens $\textbf{F}_t$ as inputs, while the queries are passed through two Temporal Decoders\cite{wang2023exploring} with the same structure as well as the corresponding detection\&tracking head and Beizer map head to decode the corresponding perception results, respectively. Refer to StreamPETR\cite{wang2023exploring}, the input and output of the Temporal Decoder can be expressed as follows
\begin{equation}
Q^c = {\rm{Temporal}}\_{\rm{Decoder}}({\textit{Query}},{\textit{Tokens}},{\textit{Memory}})
\label{eq:temporal-decoder}
\end{equation}

In SparseAD, by leveraging different inputs, \cref{eq:temporal-decoder} can be applied to both detection\&tracking and online mapping tasks. For detection\&tracking, we select the latest updated Top $K_d$ query $\mathbf{Q}^m_{dk} \in \mathcal{R}^{K_d \times C}$ from the detection memory queries $\mathbf{Q}^m_{d}$ of scene-level, along with the learnable detection query $\mathbf{Q}^i_{d}$, and the track query $\mathbf{Q}^m_{t,T-1}$ of instance-level, as the \textit{Query} input to the temporal decoder. All detection memory queries $\mathbf{Q}^m_{d}$ and sensor tokens $\textbf{F}_t$ are taken as \textit{Memory} and \textit{Tokens} input, respectively. In summary, the decoder for the detection\&tracking task can be represented as follows
\begin{equation}
\mathbf{Q}^c_d = {\rm{Temporal}}\_{\rm{Decoder}}(\text{Concat}(\mathbf{Q}^m_{dk},\mathbf{Q}^i_{d},\mathbf{Q}^m_{t,T-1}),\textbf{F}_t,\mathbf{Q}^m_{d}) 
\label{eq:det-decoder}
\end{equation}
where $\mathbf{Q}^c_d \in \mathcal{R}^{(K_d+N^i_d+N^m_i)\times C}$ is the output detection\&tracking queries of the temporal decoder in current frame. Here we omit the second dimension of $\mathbf{Q}^m_{t,T-1}$ which is $1$ for simplified descriptions.

The inputs of the decoder for the online mapping task are similar to \cref{eq:det-decoder} but simpler, because the map elements do not need to be associated and assign unique IDs between frames. Here we select the latest updated Top $K_o$ query $\mathbf{Q}^m_{ok} \in \mathcal{R}^{K_o \times C}$ from the mapping memory queries $\mathbf{Q}^m_{o}$ of scene-level, along with the learnable online mapping query $\mathbf{Q}^i_{o}$ as the \textit{Query} input to the temporal decoder. All mapping memory queries $\mathbf{Q}^m_{o}$ and sensor tokens $\textbf{F}_t$ are taken as \textit{Memory} and \textit{Tokens} input, respectively. In summary, the decoder for the online mapping task can be represented as follows
\begin{equation}
\mathbf{Q}^c_o = {\rm{Temporal}}\_{\rm{Decoder}}(\text{Concat}(\mathbf{Q}^m_{ok},\mathbf{Q}^i_{o}),\textbf{F}_t,\mathbf{Q}^m_{o}) 
\end{equation}
where $\mathbf{Q}^c_o \in \mathcal{R}^{(K_o+N^i_o)\times C}$ is the output online mapping queries of the temporal decoder in current frame.

For $\mathbf{Q}^c_d$, SparseAD uses the popular detection\&tracking head as other DETR-like detectors\cite{liu2022petr,yan2023cross,jiang2023far3d,zhang2023fully} to decode it into 3D bounding boxes which consist of position, size, yaw, and velocity. For $\mathbf{Q}^c_o$, referring to the BeMapNet\cite{qiao2023end}, SparseAD regards the sparse map elements as piece-wise Bezier curves and leverages a Bezier map head for each query to regress the Bezier curve parameters such as explicit control points, implicit control points, number of pieces, and confidence level, etc. In particular, to facilitate the downstream tasks to intuitively obtain the accurate location of the map elements, we uniformly sample $S$ points on the curve of each map element based on decoded Bezier curve parameters. Specifically, given a piece of Bezier curve with coefficients $c$, we can sample it at any point by the following

\begin{equation}
p(t)=\sum_{i=0}^n b_{i, n}(t) c_i, t \in[0,1]
\label{eq:bezier_eq_1}
\end{equation}
where
\begin{equation}
b_{i, n}(t)=\left(\begin{array}{c}
n \\
i
\end{array}\right) t^i(1-t)^{n-i}, i=0, \ldots, n
\label{eq:bezier_eq_2}
\end{equation}
where $n$ is the order of the Bezier curve. By the sampling, we can get the map curve points $\mathbf{P}^c_o\in\mathcal{R}^{(K_o+N^i_o)\times S \times 2}$ which will be utilized in downstream tasks.
\subsection{MotionPlanner}
\label{appendix-imple-motionplanner}

As shown in \cref{fig:fore-arch}, in Motion Planner, we initialize the motion query $\mathbf{Q}^i_n$ and ego query $\mathbf{Q}^i_e$ to fully utilize the priori information provided by sparse perception and temporal information from EMMB.

In Motion Planner, we consider all $K_d+N^i_d+N^m_i$ obstacles corresponding to $\mathbf{Q}^c_d \in \mathcal{R}^{(K_d+N^i_d+N^m_i)\times C}$ as instances and participate in the inference. Therefore, the number of instances in Motion Planner is $N^c_i = K_d+N^i_d+N^m_i$. Compared to the number of memory instances $N^m_i$, we set dummy history information for the newborn $ K_d+N^i_d$ instance from sparse perception. For the convenience of description, we do not distinguish between newborn instances and memory instances in the following but just replace the memory instance number $N^m_i$ with the current instance number $N^c_i$.

From EMMB, historical position information $\mathbf{P}^m_i\in\mathcal{R}^{N^c_i \times M \times 3}$ and $\mathbf{P}^m_e\in\mathcal{R}^{1 \times M \times 3}$ can be obtained for all agents, and ego-vehicle, respectively. A shared MLP is set to encode the history trajectories of both agents and ego-vehicle to obtain the history trajectory embeddings $\textbf{H}_i\in\mathcal{R}^{N^c_i \times C}$ and $\textbf{H}_e\in\mathcal{R}^{1 \times C}$ as
\begin{equation}
{{\mathbf{H}}_i} = {\rm{MLP}}({\mathbf{P}}_i^m), {{\mathbf{H}}_e} = {\rm{MLP}}({\mathbf{P}}_e^m)
\end{equation}

 For each instance, we further leverage cross-attentions to aggregate the history perception information as
\begin{equation}
{{\mathbf{H}}_{pi}} = {\rm{MHCA}}({{\mathbf{H}}_i},{\mathbf{Q}}_t^m)
\end{equation}
where MHCA denotes multi-head cross-attention\cite{vaswani2017attention}, and $\textbf{H}_{pi}\in\mathcal{R}^{N^c_i \times C}$ is the history embedding of instances. We also set a shared learnable embedding for motion query initialization while a separate learnable embedding for ego query only. In summary, motion queries and ego query in Motion Planner can be obtained by the following 
\begin{equation}
{\mathbf{Q}}_n^i = {{\mathbf{H}}_{pi}} + {{\mathbf{E}}_n},{\mathbf{Q}}_e^i = {{\mathbf{H}}_e} + {{\mathbf{E}}_e}
\end{equation}

In Motion Predictor of Motion Planner, ego query is considered equivalent to motion queries as shown in \cref{fig:fore-arch}. For ease of description, we use motion query $\mathbf{Q}_n^i$ to refer to both motion query $\mathbf{Q}_n^i$ and ego query $\mathbf{Q}_e^i$, as well as memory motion query $\mathbf{Q}_n^m$ to refer to both memory motion query $\mathbf{Q}_n^m$ and memory ego query $\mathbf{Q}_e^m$ in the interactions of Motion Predictor.

We cluster the agent future trajectories in the nuScenes \textit{train} set with the K-Means method to generate $\mathcal{K}$ shared trajectory anchors. We map them into high-dimensional feature embeddings, respectively, and combine them with motion query to form a multi-modal motion query $\mathbf{Q}_{mn}^i \in \mathcal{R}^{N^c_i\times \mathcal{K} \times C}$. However, to simplify the description, we ignore the modality of the motion query here and still use $\mathbf{Q}_{n}^i$ as the representation of the motion query.

Motion Predictor consists of N transformer layers, and each layer includes 3 types of interactions between the agents and the driving scene, including agent-memory, agent-map, and agent-social. For motion query $\mathbf{Q}_{n}^i$, its interactions with the memory motion query $\mathbf{Q}_{n}^m$, and both detection tracking query $\mathbf{Q}_{d}^c$ and online mapping query $\mathbf{Q}_{o}^c$ of the current frame can be represented as
\begin{equation}
	\mathbf{Q}_{n}^c = \text{MHCA}(\text{MHSA}(\mathbf{Q}_{n}^i),\mathbf{Q}_{n}^m / \mathbf{Q}_{d}^c/ \mathbf{Q}_{o}^c)
\end{equation}
where $\mathbf{Q}_{n}^c \in \mathcal{R}^{N^c_i \times C}$ is the updated motion query in current frame, MHSA denote multi-head self-attention\cite{vaswani2017attention}. Similarly, $\mathbf{Q}_{e}^c \in \mathcal{R}^{1 \times C}$ denotes the updated ego query in the current frame. 

For the above interaction between agents and memory motion queries, we design Temporal PE to align the predicted trajectories of the same agent across time. The motion prediction trajectory of agent $i$ for the future $L$ steps at time T can be expressed as  ${\cal F}_T^i = \{ {{\mathbf{v}}_{\mathbf{1}}},...,{{\mathbf{v}}_L}\}$, where $\mathbf{v}\in \mathcal{R}^3$ is the position in 3D space. The historical prediction trajectories in memory can then be expressed as $\left\{ {{\cal F}_{T - 1}^i,...,{\cal F}_{T - M}^i} \right\}$,  where $M$ is the max length of memory. To simplify the problem, we assume that the historical prediction trajectories have been transformed into the current ego-vehicle coordinate system. For agent $i$, the Temporal PE of memory motion query $\mathbf{E}_{n}^m$ and the Temporal PE of current motion query $\mathbf{E}_{n}^c$ can be expressed as 
\begin{equation}
 \mathbf{E}_{n}^m = \sum\limits_{k = 1}^N {{\rm{ML}}{{\rm{P}}_k}({\rm{Cat}}(\{ {{\mathbf{v}}_j} - {{\mathbf{v}}_k}|j \ge k,{\mathbf{v}} \in {\cal F}_{T - k}^i\} ))} 
 \label{eq:ts_align_pe_memory}
\end{equation}

\begin{equation}
 \mathbf{E}_{n}^c = \sum\limits_{k = 1}^N {{\rm{ML}}{{\rm{P}}_k}({\rm{Cat}}(\{ {{\mathbf{v}}_j} - {{\mathbf{v}}_0}|j \le M - k,{\mathbf{v}} \in {\cal F}_T^i\} ))} 
 \label{eq:ts_align_pe_agent}
\end{equation}

For the above interaction between agents and online mapping queries, we design Curve PE to accurately describe the location information of sparse map elements. For online mapping query $\mathbf{Q}_{o}^c$ , the Curve PE $\mathbf{E}^c_o$ can be expressed as
 \begin{equation}
     \mathbf{E}^c_o = \sum\limits_{k = 1}^J {{\rm{ML}}{{\rm{P}}}(\text{Flatten}(\mathbf{P}^c_o) )} 
 \end{equation}
where $\mathbf{P}^c_o$ is the sampling points obtained by \cref{eq:bezier_eq_1} and \cref{eq:bezier_eq_2}.

For the above interaction between agents and social,i.e., track query, we use BEV PE to describe the position information of other agents in the scenario, which is proposed in CMT\cite{yan2023cross}.

The updated motion query $\mathbf{Q}_{n}^c$ will be used to decode the multi-modal trajectory for each agent. Referring to the popular methods\cite{shi2022motion,shi2024mtr++,zhou2022hivt}, SparseAD only supervises the trajectory that closest to the ground truth in the multi-modal results during training.

{\small
\begin{algorithm}[H]
\captionsetup{font=tiny}
  \KwResult{Longitudinal and Lateral Relationship between Agents and Ego}
  $A$ \tcp*[h]{agents GT future trajectory, shaped as (F, T, N, 3)}\\
  $E$ \tcp*[h]{ego GT future trajectory, shaped as (T, N, 3)}\\
  $LoC\leftarrow None$ \tcp*[h]{initialize Longitudinal Relationship of Agents}\\
  $LaC\leftarrow None$ \tcp*[h]{initialize Lateral Relationship of Agents}\\
  $E_{speedup}$ \tcp*[h]{based on the ego GT future trajectory to generate a speedup trajectory}\\
  $E_{speeddown}$ \tcp*[h]{based on the ego GT future trajectory to generate a speed-down trajectory}\\

  \For{$i\leftarrow 0$ \KwTo \texttt{F}}{
    \For{$t\leftarrow 0$ \KwTo T\texttt{F}}{ % Corrected to use A.shape[1] for T
        \tcp{Check for Lateral Relationship with ego trajectory}
        \If{Lateral distance of $(A[i,t],E[t]) < thresh$}{
            $LaC[i] = Left$ if $A[i,t]$ is to the left of $E[t]$ else $Right$
        }
        \tcp{Check for Longitudinal Relationship with speedup trajectory}
        \If{Longitudinal distance of $(A[i,t],E_{speedup}[t]) < thresh$}{
            $LoC[i] = Front$
        }
        \tcp{Check for Longitudinal Relationship with speed-down trajectory}
        \If{Longitudinal distance of $(A[i,t],E_{speeddown}[t]) < thresh$}{
            $LoC[i] = Back$
        }
    }
  }
  \caption{Longitudinal and Lateral Relationship Generation.}
\label{alg:relationship}
\end{algorithm}}

Following the interaction between ego query and driving commands, we posit that the planned trajectories should adhere to certain scene constraints in Planning Optimizer. Consistent with previous work\cite{jiang2023vad}, we introduce the Ego-Map, Ego-Agent and Ego-Kinematic constraints. We utilize the future trajectories of both other agents and ego-vehicle to delineate their reciprocal constraints in both longitudinal and lateral directions. For further details of the Ego-Agent Relationship constraint, please refer to \cref{alg:relationship}.

Upon generating these labels, we perform a series of supervisions within the planning optimizer as outlined in \cref{sec:motion_planner}, culminating in the final planning output.

\subsection{Frame Summary}
\label{appendix-imple-summary}

After the inference of sparse perception and motion planner, the proposed SparseAD summarizes and updates the information from the multi-tasks into the EMMB to provide high-quality priori information for the future frames. We store the updated queries of each task(e.g., detection\&tracking $\mathbf{Q}^c_d$, online mapping $\mathbf{Q}^c_o$, motion $\mathbf{Q}^c_n$, and planning $\mathbf{Q}^c_e$), as well as the corresponding decoded information(e.g., 3D bounding boxes, Bezier curve control points, sampling points, multi-modal prediction trajectories, and ego-vehicle planning trajectories, etc.), into EMMB. Consistent with the common procedure,  a necessary ego transformation is conducted on all positional information in the EMMB to eliminate the influence of the motion of the ego-vehicle when the memory queries in EMMB are propagated to the next frame.

Specifically, we store the query $\mathbf{Q}^c_{dk} \in \mathcal{R}^{K_d\times C}$ with Top $K_d$ confidence level in $\mathbf{Q}^c_d$ as scene-level detection queries into EMMB. At the same time, we select high-confidence queries in $\mathbf{Q}^c_d$ according to the threshold $\mathcal{T}$  as instance-level track queries $\mathbf{Q}^c_t$, and the corresponding instances will be assigned with unique IDs. Meanwhile, downstream queries(i.e. motion queries) corresponding to track queries will be stored as instance-level memory, while the same corresponding to the detection queries without assigned IDs will be discarded.

\begin{figure}[tb]
  \centering
  \includegraphics[width=0.7\linewidth]{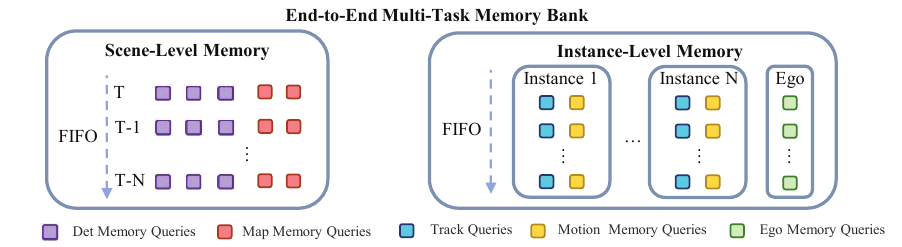}
  \caption{The schematic of End-to-End Multi-task Memory Bank(EMMB).}
  \label{fig:emmb}
\end{figure}

The End-to-End Multi-task Memory Bank (EMMB) proposed by SparseAD is shown in \cref{fig:emmb}. SparseAD utilizes two different strategies to maintain the scene-level and instance-level memory in the EMMB. For scene-level memory, we store the memory as a frame unit and follow the FIFO(First In First Out) principle. For actual streaming data, it can be assumed that the scene-level memory always stores the perceptual information of the previous $M$ frames. In contrast, instance-level stores memory as instance units and follows the FIFO principle for the memory belonging to the same instance, while the instance itself has no lifetime limit. When a new obstacle is detected by the sparse perception module, a new instance is added to the instance-level memory for storing its corresponding information. When the instance is not detected for consecutive frames (e.g., disappearances, miss detection, or ID switches), the instance with all the corresponding historical information contained will be removed. Two-level memory design allows SparseAD to fully utilize both global (scenario) and local (instance) temporal information to achieve powerful end-to-end autonomous driving performance.

\section{Experiments}
\label{appendix-exps}
\subsection{Datasets}
We conduct experiments on the challenging nuScenes \cite{caesar2020nuscenes} dataset, which contains 700/150/150 scenes for the training/validation/testing set respectively. Each scene has a duration of around 20 seconds and contains roughly 40 key-frames annotated at 2Hz, where each sample includes 6 images captured by 6 cameras covering 360° FOV horizontally as well as point clouds collected by both LiDAR and radar sensors.
\subsection{Configuration Details}

Schematically, SparseAD consists of three parts, including sensor encoders, sparse perception and motion planner. Specifically, we have two variants of \model, where \model-Base and \model-Large adopt VoVNetv2\cite{lee2020centermask} and ViT-Large\cite{dosovitskiy2020image} as image backbone with the input image size of $1600\times640$ and $1600\times800$, respectively.

For the nuScenes dataset, the range of SparseAD is set to $\left[-51.2m, 51.2m\right]$ for both the X and Y axes. All sub-tasks in SparseAD are conducted and evaluated in this range, including detection, tracking, online mapping, motion prediction, and planning. In SparseAD, we set the maximum history length of EMMB $M$ to 5, which corresponds to about 2.5s of historical temporal information.

\subsubsection{Sparse Perception.}

For the detection\&tracking task, we set 644 learnable detection queries in each frame, and 256 sence-level detection memory queries from EMMB, following StreamPETR\cite{wang2023exploring}, as well as the propagated high-score queries from the last frame as track queries. For each track query, we set a drop probability of 0.5 and introduce an approximate negative sample with a probability of 0.2 during training to balance the supervision of instance-level track queries and the scene-level detection queries. For the online mapping task. We use $\left[35, 30, 15, 25\right]$ learnable online mapping queries, $\left[14, 12, 6,10\right]$ scene-level mapping memory query for \textit{Divider}, \textit{Cross}, \textit{Road Segment}, and \textit{Lane}, respectively, referring to BeMapNet\cite{qiao2023end}. The setting of piece-wise Bezier curve $\langle \text{max pieces}, \text{max degree}\rangle$ for
\textit{Divider}, \textit{Cross}, \textit{Road Segment}, and \textit{Lane} are set to
$ \langle3, 2\rangle, \langle1, 1\rangle, \langle7, 3\rangle$, and $ \langle7, 3\rangle$.
\subsubsection{Motion Planner.}

The module utilizes the temporal information of the last 5 frames (we set $M=5$) to predict the future motion of other agents for the next 12 frames (6s) while planning the motion of the ego car for the next 6 frames (3s). We conduct a K-Means clustering algorithm on the nuScenes \texttt{train} set to obtain the prior intention points and trajectories for motion queries. For each agent (including ego-vehicle), we set 6 motion queries to predict the multi-modal future trajectories. We only supervise the predicted trajectory closest to the ground truth trajectory during training. The multi-modal motion prediction of the ego-vehicle is input to the Planning Optimizer as strong priori information. Finally, SparseAD outputs a high-quality, safe, and controllable planning trajectory for the ego-vehicle.

\subsubsection{Training and Inference.}
\label{sec:train_infer_detail}

SparseAD can be trained and tested online. The training is divided into three stages in total. In \textbf{stage 1}, the sparse perception module is trained from scratch for the basic perception capability of the driving scene, including detection and online mapping. In \textbf{stage 2}, we introduce the two-level memory to optimize the tracking performance of the sparse perception module. In \textbf{stage 3}, we keep the backbone and the sparse perception module frozen to train the motion planner module. In particular, \textbf{stage 1 and stage 2 can be merged} during training for a shorter training time with slight performance degradation in exchange, which is a trade-off. In our experiments, we use the AdamW\cite{loshchilov2017decoupled} optimizer and Cosine Annealing\cite{loshchilov2016sgdr} scheduler to train the model on 8 NVIDIA A100 80GB GPUs. Especially, as shown in \cref{tab:joint-results}, SparseAD-Base only needs a max of \textbf{17.5GB} GPU memory during the entire training process, which means it's supported to train on \textbf{NVIDIA RTX 3090 24GB GPUs} during the entire process. The ablation experiments in \cref{sec:experiments} are all conducted on NVIDIA RTX 3090 24GB GPUs, where the input size of the images is set to $800\times320$. During inference, when an agent is not detected by the model for 8 consecutive frames (4s), the corresponding instance-level memory will be removed to keep a trade-off between performance and efficiency.

\subsection{Metrics}
\subsubsection{Metrics for Sparse Perception.}

To maintain consistency with other popular methods, we adopt the commonly used metrics for the evaluation of perception tasks on the nuScenes \cite{caesar2020nuscenes} dataset. Specifically, we use \textbf{NDS} (nuScenes detection score), \textbf{mAP} (mean Average Precision) as the main metrics to assess the detection performance of \model. As the composite index of detection performance provided by the nuScenes dataset itself, NDS takes the indicators from different dimensions into account including \textbf{mAP}, \textbf{mATE} (mean Average Translation Error), \textbf{mAOE} (mean Average Orientation Error), \textbf{mASE} (mean Average Scale Error), \textbf{mAVE} (mean Average Velocity Error), and \textbf{mAAE} (mean Average Attribute Error). For the task of tracking, we adopt metrics such as \textbf{AMOTA} (Average Multi-Object Tracking Accuracy), \textbf{AMOTP} (Average Multi-Object Tracking Precision), \textbf{RECALL} and \textbf{IDS} (ID-Switch) for evaluation, where AMOTA and AMOTP are the average results of \textbf{MOTA} and \textbf{MOTP} at different recall rates respectively. The specific calculation methods can be referred to in \cite{caesar2020nuscenes}. As for online mapping, we categorize map elements into four types including Divider, Cross, Road Segment, and Lane, and we calculate the \textbf{AP} (Average Precision) of each type with the TP threshold of $[0.5m, 1m, 2m]$ and ultimately obtain the \textbf{mAP} (mean Average Precision) for each category as the evaluation metric.

\subsubsection{Metrics for Motion Planner.}
Being consistent with mainstream methods of motion prediction, we employ metrics such as \textbf{minADE} (minimum Average Distance Error) and \textbf{minFDE} (minimum Final Distance Error) as the main metrics to evaluate the prediction performance of SparseAD. In addition, we also use \textbf{MR} (Miss Rate) and \textbf{EPA} (End-to-end Prediction
Accuracy) proposed in \cite{gu2023vip3d} to assess the quality of end-to-end motion prediction, which helps to reflect the performance more comprehensively since the metrics of minADE and minFDE are calculated only on TPs. Our benchmark is consistent with VIP3D \cite{gu2023vip3d} to ensure a fair comparison. As for the planning, we utilize the \textbf{L2} error and \textbf{ColRate} (Collision Rate) widely-used in previous methods \cite{hu2023planning, jiang2023vad} to demonstrate the performance of planning for the ego-vehicle. Concretely, the L2 error represents the fitting degree between the planned and actual trajectories of the ego-vehicle, and the ColRate reflects the probability of collision between the ego-vehicle and surrounding agents. It should be pointed out that we have re-examined the calculation method of collision rate in other methods and adopted a sparse way to calculate the collision rate more accurately. The detail will be introduced in \cref{appendix-discussion}.

\subsection{Supplementary Experiments}
\begin{table}[t]
\centering
\caption{Results of 3D detection and multi-object tracking on nuScenes \texttt{test} dataset.}
\label{tab:det_tracking_test_set}
\tiny
\resizebox{\textwidth}{!}{
% {
\setlength{\tabcolsep}{3.5pt}
\begin{tabular}{l|c|c|ccc|l|cccc} 
\toprule
\textbf{Methods}&  \textbf{mAP}$\uparrow$  &\textbf{NDS}$\uparrow$  & \textbf{mATE}$\downarrow$  &\textbf{mAOE}$\downarrow$   &\textbf{mAVE}$\downarrow$ & \textbf{Methods}& \textbf{AMOTA}$\uparrow$  & \textbf{AMOTP}$\downarrow$ & \textbf{RECALL}$\uparrow$ & \textbf{IDS}$\downarrow$\\
\midrule 
\multicolumn{5}{l}{\textit{Detection-only methods}}&&\multicolumn{5}{l}{\textit{Tracking-only methods}}\\
\midrule
% ResNet101
StreamPETR\cite{wang2023exploring}   &0.546& 0.633 &0.485&  \underline{0.307}& 0.250                  &DEFT\cite{chaabane2021deft} &0.177 &1.564 &0.338&  6901\\ 
PETRv2\cite{liu2023petrv2}         &0.490& 0.582& 0.561& 0.361 &0.343               &QD3DT\cite{hu2022monocular} &0.217& 1.550 &0.375  &6856\\ % PETR Table 1
BEVFormer \cite{li2022bevformer}    & 0.481& 0.569& 0.582&  0.375& 0.378            & CC-3DT$^*$~\cite{fischer2022cc} &0.410 &1.274 &0.538&3334\\ % BEVFormer Table 4.
Sparse4D\cite{lin2022sparse4d}  &0.511 & 0.595&0.533&  0.369 &0.317                 &QTrack\cite{yang2022quality}& 0.480 &1.107& 0.569&  1484\\ % PolarDETR Table 1
SRCN3D\cite{shi2022srcn3d}&0.396 &0.463&0.673 & 0.403& 0.875& SRCN3D\cite{shi2022srcn3d}& 0.398 &1.317 &0.538&-\\
DORT\cite{qing2023dort}& 0.545& 0.625&\underline{0.440} & 0.370& 0.250 &DORT\cite{qing2023dort}&\underline{ 0.576} &\underline{0.951}&-&-\\
\midrule
\multicolumn{7}{l}{\textit{End-to-End Multi-Object Tracking Methods}}\\
\midrule
MUTR3D\cite{zhang2022mutr3d}  &-&-&-&-&-& MUTR3D\cite{zhang2022mutr3d} &0.270& 1.494 &0.411& 6018\\
PF-Track\cite{pang2023standing} &-&-&-&-&- &PF-Track\cite{pang2023standing} &0.434 &1.252 & 0.538&  \textbf{249}\\
DQTrack\cite{li2023end}&-&-&-&-& -& DQTrack\cite{li2023end} &0.523& 1.096 &  0.622 &1204\\
Sparse4Dv3\cite{lin2023sparse4d} &\textbf{0.570}& \textbf{0.656}&\textbf{0.412} & 0.312& \textbf{0.210} & Sparse4Dv3\cite{lin2023sparse4d} &0.574& 0.970  &\underline{0.669}  &669\\
\midrule
\multicolumn{7}{l}{\textit{End-to-End Autonomous Driving Methods}}\\
\midrule
% UniAD\cite{hu2023planning} &-&-&-	&	-	&-& UniAD\cite{hu2023planning} &- &- &- &-\\ 
\rowcolor[gray]{.9}  \model-B& 0.514 & 0.605 &0.505 &0.337 &0.308	& \model-B &0.502 &1.065 &	0.643 &349 \\ 
\rowcolor[gray]{.9}  \model-L& \underline{0.566} & \underline{0.648} &0.454 &\textbf{0.288} &\underline{0.237}	& \model-L &\textbf{0.588} &\textbf{0.904} &	\textbf{0.691} &\underline{344} \\ %这里使用的是大分辨率5帧3梯度训出来的临时结果，后续还会更改
\bottomrule
\end{tabular}}
\end{table}

\subsubsection{Evaluation on Test Set.}
Although the experiments in \cref{sec:experiments} have evaluated and the comprehensive performance of SparseAD and compared it with other methods in detail on the nuScenes \texttt{val} set, we also report the performance of SparseAD on the nuScenes \texttt{test} set to further validate the perception performance. As shown in \cref{tab:det_tracking_test_set}, \model-Base achieves comparable performance compared to popular methods \cite{liu2023petrv2, li2022bevformer, lin2022sparse4d, shi2022srcn3d} but is still inferior to the state-of-the-art single-task methods \cite{lin2023sparse4d, wang2023exploring}. But in the meanwhile, \model-Large reaches \textbf{56.6\%} mAP, \textbf{64.8\%} NDS, and \textbf{58.8\%} AMOTA, even slightly better than the performance of state-of-the-art method\cite{lin2023sparse4d}, which shows the great potential of the end-to-end paradigms for autonomous driving paradigm. It is worth noting that UniAD\cite{hu2023planning} have not been evaluated on the nuScenes \texttt{test} set, so there is no methods that can be completely fairly compared with ours. As an end-to-end method, only the results of detection and tracking can be evaluated on the nuScenes official benchmark that can be fairly compared to other methods. Therefore, we can't provide evaluation results for other tasks (e.g., mapping, motion prediction, and planning) on the nuScenes \texttt{test} set.

\begin{table*}[t]
	\caption{A comprehensive comparison of the performance of SparseAD-B with or without the end-to-end gradient optimization in stage 3 training. B.: Backbone, P.: Sparse Perception, M.: Motion Planner, T.Mem: Training memory footprint.}
	\begin{center}
		\tiny
		\vspace{-10pt}
		\resizebox{1\textwidth}{!}
		{
			\begin{tabular}{ccc|cc|cc|ccc|cccc|cccc|c}
				\toprule
				\multicolumn{3}{c|}{Gradient} &
				\multicolumn{2}{c|}{Detection} & 
				\multicolumn{2}{c|}{Tracking} & 
				\multicolumn{3}{c|}{Motion Prediction} & 
				\multicolumn{4}{c|}{Planning L2$\downarrow$} &
				\multicolumn{4}{c|}{Planning Col.$\downarrow$} &
				Efficiency \\
				B.&P.&M.& mAP$\uparrow$ & NDS$\uparrow$ & AMOTA$\uparrow$&Recall$\uparrow$& minADE$\downarrow$&minFDE$\downarrow$&MR$\downarrow$ &1s &2s &3s &avg.&1s&2s&3s &avg.&  T. Mem \\
				\midrule
				&&\Checkmark& 0.475& 0.578& 0.530& 0.608 & 0.83& 1.58& 0.187&0.15&0.31&0.57& 0.35&0.00&0.06&0.21& 0.09&4.0\\
				\Checkmark&\Checkmark&\Checkmark&0.455& 0.559& 0.521 & 0.622&0.83&	1.58&	0.180&0.15&0.31&0.56& 0.34 &0.00&0.04&0.15&0.08&18.1\\
				\midrule
				\multicolumn{3}{c|}{\textit{Difference}} &\textcolor{blue}{-0.020} & \textcolor{blue}{-0.019}& \textcolor{blue}{-0.009} & \textcolor{red}{+0.014} & 0.00 & 0.00 & \textcolor{red}{-0.007}& 0.00&0.00&\textcolor{red}{-0.01}&\textcolor{red}{-0.01}&0.00&\textcolor{red}{-0.02}&\textcolor{red}{-0.06}&\textcolor{red}{-0.01}& \textcolor{blue}{+12.1} \\

				\bottomrule
			\end{tabular}% }
		}
	\end{center}
	% \vspace{2pt}
	% \vspace{-20pt}
	\label{tab:joint-results-gradient}
\end{table*}

\subsubsection{End-to-End Gradient Optimization.} Notice that during the training process in \textbf{stage 3}, SparseAD does not turn on the gradient of the entire model for training. However, due to the efficient design of SparseAD, it is feasible to achieve end-to-end gradient optimization for the whole model. Based on SparseAD-Base, we evaluated the influence of the different gradient settings on the performance in the stage 3 training phase. As shown in \cref{tab:joint-results-gradient}, the sub-task performance of motion prediction and planning is slightly better with the gradients from Motion Planner propagated to upstream perception tasks and the extra memory footprint caused by end-to-end optimization is also acceptable for academic-level computation power (e.g., NVIDIA RTX 3090 GPU). However, the end-to-end optimization also slightly degrades the perception capabilities of SparseAD. Considering that the accuracy and safety of ego-vehicle planning is the ultimate goal of an end-to-end autonomous driving method, a slight performance degradation in perception is acceptable here. Overall, we believe that the end-to-end optimization in stage 3 can be regarded as a kind of trade-off between perception and planning performance.

\subsection{Runtime Analysis}

\begin{table*}[t]
	\caption{Module runtime. The inference speed is measured for SparseAD-Base and SparseAD-Large on one NVIDIA RTX A100 80GB GPU.}
	\begin{center}
		\tiny
		\vspace{-10pt}
		{
			\begin{tabular}{p{4cm}|p{2cm}<{\centering}p{2cm}<{\centering}|p{2cm}<{\centering}p{2cm}<{\centering}}
				\toprule
				\multirow{2}{*}{Module}&
				% \multicolumn{2}{c|}{VoVNet-V2-99\cite{lee2020centermask}} &
				% \multicolumn{2}{c}{ViT-Large\cite{dosovitskiy2020image}}  \\
                \multicolumn{2}{c|}{SparseAD-B} &
				\multicolumn{2}{c}{SparseAD-L}  \\
				& Latency(ms) & Proportion & Latency(ms) & Proportion\\
				\midrule
				Backbone& 85& 29.6\%&1153&83.1\%\\
				Sparse Perception&104& 36.2\%&130&9.4\%\\
				Motion Planner&75&26.1\%&75&5.4\%\\
				Output\&Memory Update&23 &8.1\%&28 & 2.1\%\\
				\midrule
				\textit{Total}&287 & 100.0\% & 1386 & 100\%\\
				\bottomrule
			\end{tabular}% }
		}
	\end{center}
	\label{tab:runtime}
\end{table*}

 We evaluate the running time of each module of SparseAD-Base and SparseAD-Large and the results are shown in \cref{tab:runtime}. The inference latency of SparseAD-Base is \textbf{287ms}, of which sparse perception is the most time-consuming module with 36.2\% of total latency. Notice that the runtime for maintaining EMMB in SparseAD is 23ms, which is only about 8.1\% of the total latency, allowing us to efficiently utilize the long temporal information. On the other hand, the inference latency of SparseAD-Large is \textbf{1386ms}, where the massive ViT-Large backbone becomes the main component of inference latency, which occupies 83.1\% of the total inference latency.  SparseAD-Large has a comprehensive performance boost compared with SparseAD-Base, with only a slight increase in latency in terms of Sparse Perception, Motion Planner, and Output \& memory update. These results also demonstrate that the design of SparseAD can efficiently benefit from an advanced and large-scale backbone.

% \clearpage
\section{Variant Paradigm of SparseAD}
\label{appendix-variant}
\subsection{Motivation And Method}

\subsubsection{Motivation.}

As mentioned in \cref{sec:sparse_perception}, there is an apparent supervision imbalance between detection and track queries which results in a non-negligible degradation of detection performance, and also affects the performance of tracking. In order to alleviate this problem, we revisited the entire framework of Sparse Perception and realized that the scene-level detection memory queries, which also originate from historical frames, to some extent, may be able to act as track queries. Moreover, during the training phase, the scene-level detection memory queries are not forcibly bound to the ground truth through ID relationships, and thus do not suffer from the problem of supervision imbalance. In other words, we may be able to achieve end-to-end tracking using only scene-level detection memory queries without causing a decline in detection performance.

\subsubsection{Method.}

During the training phase, we discarded the track queries from the instance-level memory,which allowed us to omit the \textbf{stage 2} mentioned in \cref{sec:train_infer_detail}. At the same time, we adjusted the ratio of learnable detection queries to scene-level detection memory queries from the original 644:256 to 200:700, which helps ensure the consistency of tracking. In order to replace the track queries and provide stable historical information of obstacles for downstream tasks, we concatenate detection queries that match the same ground truth to form a new instance. That is to say, the historical information of an instance may come from different detection queries. During the inference phase, the bounding boxes corresponding to the scene-level detection memory queries will inherit the IDs from the previous frame, while the bounding boxes corresponding to the learnable detection queries will be assigned with new IDs. The same with in the training phase, we concatenate queries with the same ID to form a new instance for downstream tasks. The other details of this paradigm remain consistent with the descriptions in \cref{sec:method} and \cref{appendix-imple}.

\subsection{Validation Experiments}
We conduct experiments on nuScenes val set. The performance of multiple driving tasks are shown in \cref{tab:hard-asso-det-tracking}. As we expected, \model without hard association achieves significant performance gains. Compared to SparseAD-Base, there is an increase of $+1.2\%$ in NDS, $+2.4\%$ in mAP, and $+1.7\%$ in AMOTA. With ViT-Large as the image backbone, the increases are $+2.3\%$ in NDS, $+3.7\%$ in mAP, and $+2.3\%$ in AMOTA. However, it should be noted that due to the lack of hard association during training, there is also a noticeable increase in IDS. We also compared the prediction and planning performance of the two methods, with the results shown in \cref{tab:hard-asso-motion-planner}. Although \model without hard association appears to perform better in perception according to certain metrics, the prediction and planning capabilities of both methods are comparable. We believe that even though \model without hard association recalls more challenging samples, the excessive IDS means that the trajectory consistency is not good enough, which ultimately leads to similar prediction and planning performance.

\begin{table*}[t]
	\caption{Comparative experiments on the effect of hard association supervision during training on the sparse perception performance of SparseAD.}
	\begin{center}
		\tiny
		\vspace{-10pt}
		\resizebox{1\textwidth}{!}
		{
			\begin{tabular}{c|c|cc|ccccc|ccccc|c}
				\toprule
				\multirow{2}{*}{Backbone}& \multirow{2}{*}{Hard-Asso.}& \multicolumn{7}{c|}{Detection} & \multicolumn{5}{c}{Tracking}& Mapping\\
				&&NDS$\uparrow$&	mAP$\uparrow$&	ATE$\downarrow$&	ASE$\downarrow$&	AOE$\downarrow$	&AVE$\downarrow$&	AAE$\downarrow$& AMOTA$\uparrow$&	AMOTP$\downarrow$&	Recall$\uparrow$ & IDs$\downarrow$ &Frag$\downarrow$& mAP$\uparrow$\\
				\midrule
				\multirow{2}{*}{V2-99\cite{lee2020centermask}}& \Checkmark& 0.579&	0.475&	0.556&	0.264&	0.295&	0.288&	0.186&0.530&	1.087&	0.608 &297&	454& 34.2\\
				&\XSolidBrush &0.591&	0.499&	0.554&	0.260&	0.323&	0.256&	0.198& 0.547&	1.115&	0.629 &639&	658	& 33.5\\
				\midrule
				\multicolumn{2}{|c|}{\textit{Difference}}& \textcolor{red}{+0.012}& \textcolor{red}{+0.024} &\textcolor{red}{-0.002} &\textcolor{red}{-0.004}& \textcolor{blue}{+0.028} &\textcolor{red}{-0.032}& \textcolor{blue}{+0.008}& \textcolor{red}{+0.017} &\textcolor{blue}{+0.028}& \textcolor{red}{+0.021}& \textcolor{blue}{+442}& \textcolor{blue}{+204}&\textcolor{blue}{-0.7}\\
				\midrule
				\multirow{2}{*}{ViT-L\cite{dosovitskiy2020image}}& \Checkmark& 0.625&	0.536&	0.531&	0.252&	0.219&	0.230&	0.199& 0.606&	0.933&	0.706&251&	335	& 34.6\\
				&\XSolidBrush&0.648&	0.573&	0.500&	0.251&	0.223&	0.210&	0.202&0.629&	0.942&	0.733 &958&	507 & 34.3\\
				\midrule
				\multicolumn{2}{|c|}{\textit{Difference}} & \textcolor{red}{+0.023}& \textcolor{red}{+0.037}& \textcolor{red}{-0.031}& \textcolor{red}{-0.001}& \textcolor{blue}{+0.004}& \textcolor{red}{-0.020}& \textcolor{blue}{+0.003}& \textcolor{red}{+0.023}& \textcolor{blue}{+0.009}& \textcolor{red}{+0.027}& \textcolor{blue}{+707}& \textcolor{blue}{+172} & \textcolor{blue}{-0.3}\\
				\bottomrule
			\end{tabular}% }
		}
	\end{center}
	\label{tab:hard-asso-det-tracking}
\end{table*}

\begin{table*}[t]
	\caption{Comparative experiments on the effect of hard association supervision during training on the motion planner performance of SparseAD.}
	\begin{center}
		\tiny
		\vspace{-10pt}
		{
			\begin{tabular}{c|c|ccc|cccc|cccc}
				\toprule
				\multirow{2}{*}{Backbone}& \multirow{2}{*}{Hard Asso.}& \multicolumn{3}{c|}{Motion Prediction} & \multicolumn{4}{c|}{Planning L2(m)$\downarrow$} & \multicolumn{4}{c}{Planning Col.$\downarrow$}\\
				&&minADE$\downarrow$&	minFDE$\downarrow$&	MR$\downarrow$&1s &2s&3s&avg.&1s &2s&3s&avg.\\
				\midrule
				\multirow{2}{*}{V2-99\cite{lee2020centermask}}& \Checkmark& 0.830&	1.584&	0.187& 0.15&0.31&0.57&0.35 &	0.00&0.06&0.21&0.09\\
				&\XSolidBrush & 0.823&	1.551&	0.191&	0.15&0.31&0.56&0.34&	0.00&0.04&0.23& 0.09\\
				\midrule
				\multicolumn{2}{|c|}{\textit{Difference}}& \textcolor{blue}{+0.003} &\textcolor{red}{-0.033}& \textcolor{blue}{+0.004}& 0.00& 0.00& \textcolor{red}{-0.01}& \textcolor{red}{-0.01}& 0.00 &\textcolor{red}{-0.02}& \textcolor{blue}{+0.02}& 0.00\\
				\midrule
				\multirow{2}{*}{ViT-L\cite{dosovitskiy2020image}}& \Checkmark&0.741&	1.458&	0.169& {0.15}&{0.31}&{0.56}&{0.34}&	{0.00}&{0.04}&{0.15}&{0.06}\\
				&\XSolidBrush& 0.740&	1.453&	0.173&	0.15&0.31&0.56&0.34&	0.00&0.04&0.14& 0.06\\
				\midrule
				\multicolumn{2}{|c|}{\textit{Difference}} & \textcolor{red}{-0.001}& \textcolor{red}{-0.005}& \textcolor{blue}{+0.004}& 0.00& 0.00 &0.00& 0.00& 0.00& 0.00& \textcolor{red}{-0.01} &0.00\\
				\bottomrule
			\end{tabular}% }
		}
	\end{center}
	\label{tab:hard-asso-motion-planner}
\end{table*}

\section{Discussion about the Collision Rate Metric}
\label{appendix-discussion}
\begin{table*}[t]
	\caption{Comparison of the collision rate calculated by two different calculation methods (i.e. Occupancy-Based\cite{hu2022st} and Sparse-Based) of the ground truth future trajectory of the ego-vehicle.}
	\begin{center}
		\tiny
		\vspace{-10pt}
%		\resizebox{1\textwidth}{!}
		{
			\begin{tabular}{p{3cm}|p{1.5cm}<{\centering}|p{0.7cm}<{\centering}p{0.7cm}<{\centering}p{0.7cm}<{\centering}p{0.7cm}<{\centering}}
				\toprule
				\multirow{2}{*}{Metirc}& \multirow{2}{*}{Resolution}& \multicolumn{4}{c}{Collision(\%)} \\
				&& 1s& 2s& 3s& avg.\\
				\midrule
				Occupancy-Based\cite{hu2022st}&0.5m& 10.39	&10.33&	10.25& 10.32\\
				Sparse-Based(Ours)&-&0.58&	0.58&	0.58& 0.58\\
				\bottomrule
			\end{tabular}% }
		}
	\end{center}
	% \vspace{2pt}
	% \vspace{-20pt}
	\label{tab:gt-collision}
\end{table*}

\begin{table*}[t]
	\caption{Comparison of the collision rate evaluated on the planning results of different methods by the proposed sparse-based collision metric. $\ddag$: the results are reproduced using the official checkpoint.}
	\begin{center}
		\tiny
		\vspace{-10pt}
		%		\resizebox{1\textwidth}{!}
		{
			\begin{tabular}{c|c|cccc|cccc}
				\toprule
				\multirow{2}{*}{Metirc}& \multirow{2}{*}{Col. Optimizer}& \multicolumn{4}{c|}{L2(m)}& \multicolumn{4}{c}{Collision(\%)} \\
				&&1s& 2s& 3s& avg.& 1s& 2s& 3s& avg.\\
				\midrule
				\multirow{2}{*}{UniAD$\ddag$\cite{hu2023planning}}&\Checkmark&0.44&0.69&1.03 &0.72&1.07&	1.35&	1.62&1.35\\
				&\XSolidBrush&0.31&0.57&0.93&0.60&0.52&	0.73&	1.19&0.81\\
				\midrule
				{SparseAD-B}&-&0.15&0.31&0.56&0.29&	0.00& 0.04&	0.14&0.06\\
				\bottomrule
			\end{tabular}% }
		}
	\end{center}
	% \vspace{2pt}
	% \vspace{-20pt}
	\label{tab:pred-collision}
\end{table*}
\begin{figure}[tb]
  \centering
  \begin{subfigure}{0.6\linewidth}
  \includegraphics[width=1\linewidth]{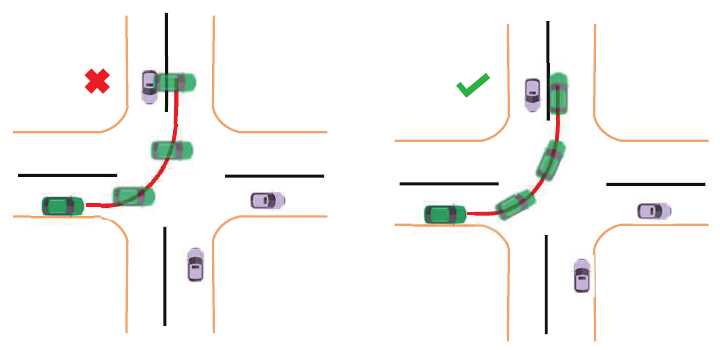}
  \caption{}
  \label{fig:coll-rot}
  \end{subfigure}
    \begin{subfigure}{0.3\linewidth}
  \includegraphics[width=1\linewidth]{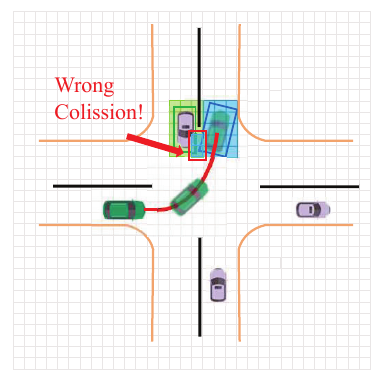}
  \caption{}
  \label{fig:coll-grid}
  \end{subfigure}
  \caption{Illustration of the disadvantages of traditional calculation methods for collision rate. (a) Ignoring the variance in the yaw of ego-vehicle could lead to potentially incorrect collision judgments. (b) The limited resolution of the occupancy grid may also lead to a incorrect collision judgment.}
  \label{fig:collision-metric}
\end{figure}
The collision rate is a crucial metric to evaluate the safety of planning trajectories in end-to-end autonomous driving methods. However, we argue that there are \textbf{two significant disadvantages} of the commonly used calculation \cite{hu2022st} for the collision rate which results in unreliable evaluation. \textbf{Firstly}, as previously highlighted\cite{li2023ego}, traditional metrics do not take variations in ego-vehicle heading into account, particularly in turning scenarios where the vehicle's bounding box orientation improperly remains as if it were moving straight. We illustrate this type of case in \cref{fig:coll-rot}. \textbf{Secondly}, the occupancy(OCC) grid assessments which are commonly used to calculate the collision status by the previous studies only compromise about 0.5 meters in grid resolution for a trade-off between efficiency and accuracy. However, this reduction in accuracy could result in the wrong judgment of collision between the ego-vehicle and surrounding agents in evaluations since 0.5 meters is far larger than the possible minimum safe distance between the ego-vehicle and other agents in complex scenarios, which may also result in unreliable evaluation, as shown in \cref{fig:coll-grid}.

Because of possible imprecision of IMU, the position of the ego-vehicle may be slightly biased from the actual position and make a "collision" with the other agents in the driving scenario. In previous calculation of collision rate, to ensure no collision for the ground truth trajectory of the ego-vehicle, the "collision" cases of ground truth are excluded from benchmarks, leaving them not involved in the evaluation. However, we find that the "collision" case of ground truth is significantly overrated due to the insufficient resolution of the occupancy grid and the neglect of the variants in the yaw of the ego-vehicle. As shown in \cref{tab:gt-collision}, the "collision" rate of the ego-vehicle's ground truth trajectory is over \textbf{10\%} which is \textbf{impossible} and \textbf{unacceptable} in real driving during data collection even taking the error of IMU into account.

Actually, we argue that most of the misjudged "collision" cases are the \textbf{hard-case}, which is the critical key to a comprehensive evaluation for planning. Unfortunately, these hard cases are excluded by the evaluation of planning performance in previous end-to-end autonomous driving methods. To address this problem, we proposed a sparse-based method to calculate the real and precise collision status for the reliable evaluation of planning capabilities of end-to-end autonomous driving methods. Concretely, when calculating collisions for each frame, obstacles' positions and their headings at any given moment are derived from the nuScenes dataset, and the ego-vehicle's heading is approximated by comparing positions between the current and preceding frames of the planned trajectory. This approach removes the reliance on occupancy grids, with collision rates being computed only when overlaps occur between the bounding box of the ego-vehicle and those of other obstacles.

As shown in \cref{tab:gt-collision}, after the calculation by the proposed metric, the collision rate of ground truth trajectories of ego vehicles is only \textbf{0.58\%}, which is more reasonable and nearly 20x lower than the results calculated by coarse occupancy grids. The differences between the proposed sparse-based method and the traditional occupancy-based method are illustrated \cref{fig:coll-rot}.

Due to the considerations of fair comparison to the state-of-the-art end-to-end method and fully validating the effectiveness of the proposed sparse-based collision metric, we re-evaluate UniAD under the proposed metrics and compare it with SparseAD-B, as shown in \cref{tab:pred-collision}. The results show a significant collision rate degradation on UniAD\cite{hu2023planning} compared with the results in its paper which meets expectations due to the introduction of the hard case.

We hope that the discussion in this section and the proposed sparse-base collision metric may inspire and help future research on end-to-end autonomous driving methods.

\clearpage
\section{Qualitative results on Various Cases}
\label{appendix-vis}
\begin{figure}[h]
  \centering
  \includegraphics[width=1\linewidth]{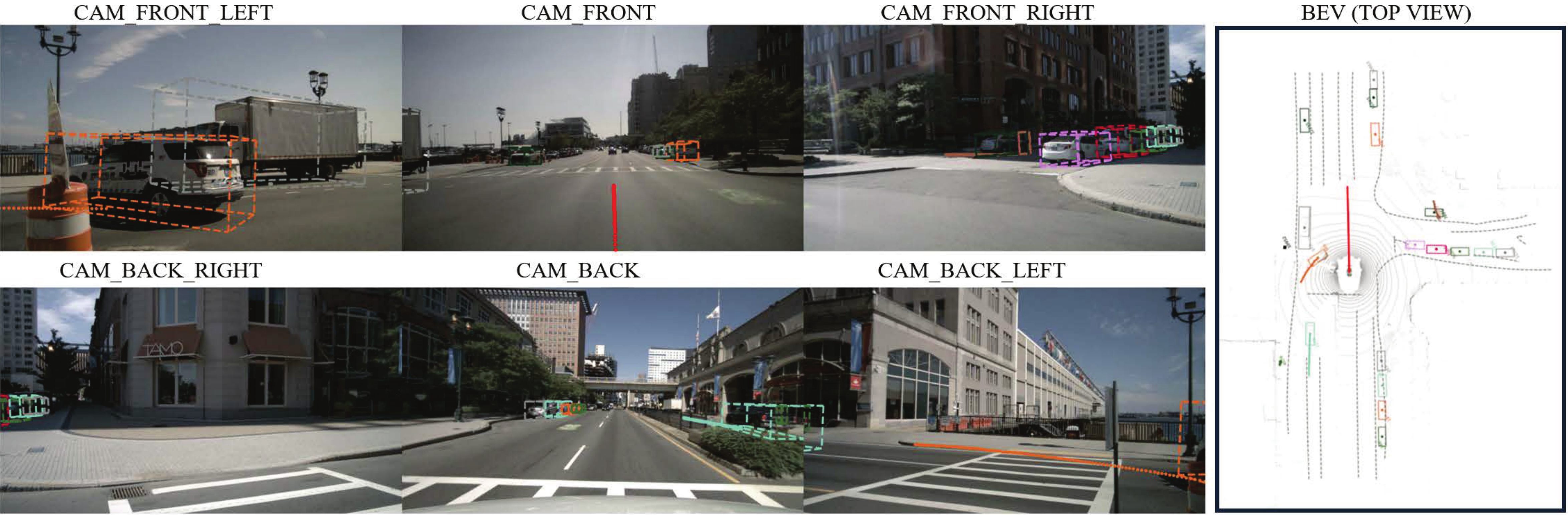}
  \includegraphics[width=1\linewidth]{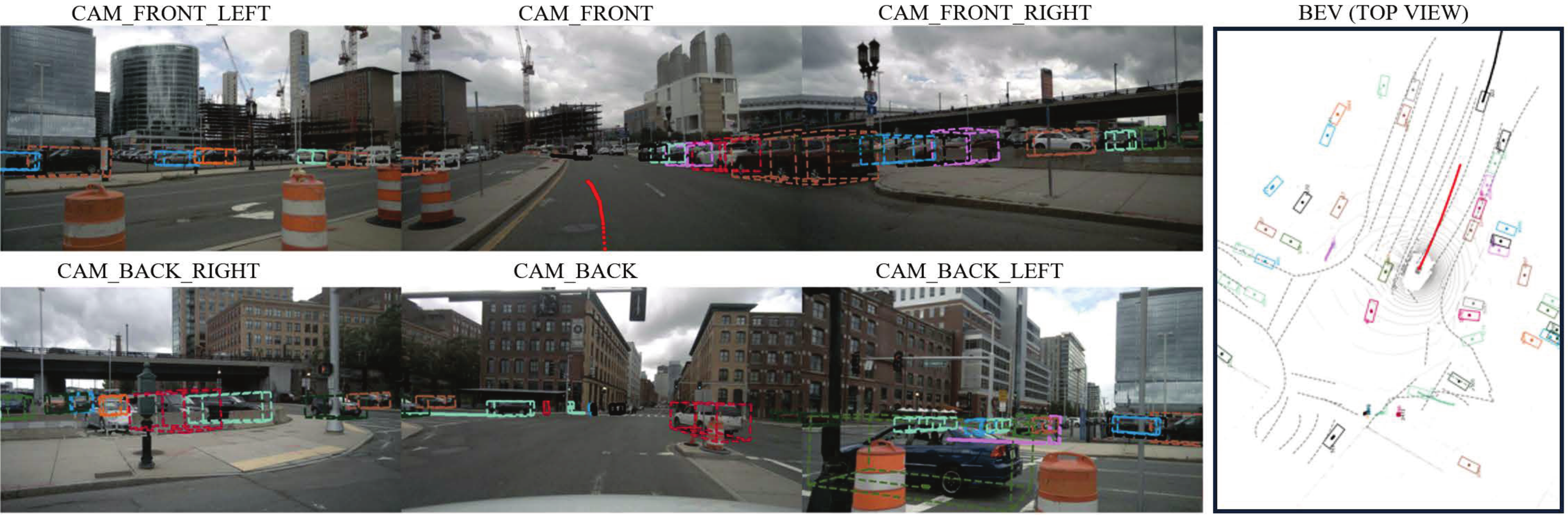}
  \includegraphics[width=1\linewidth]{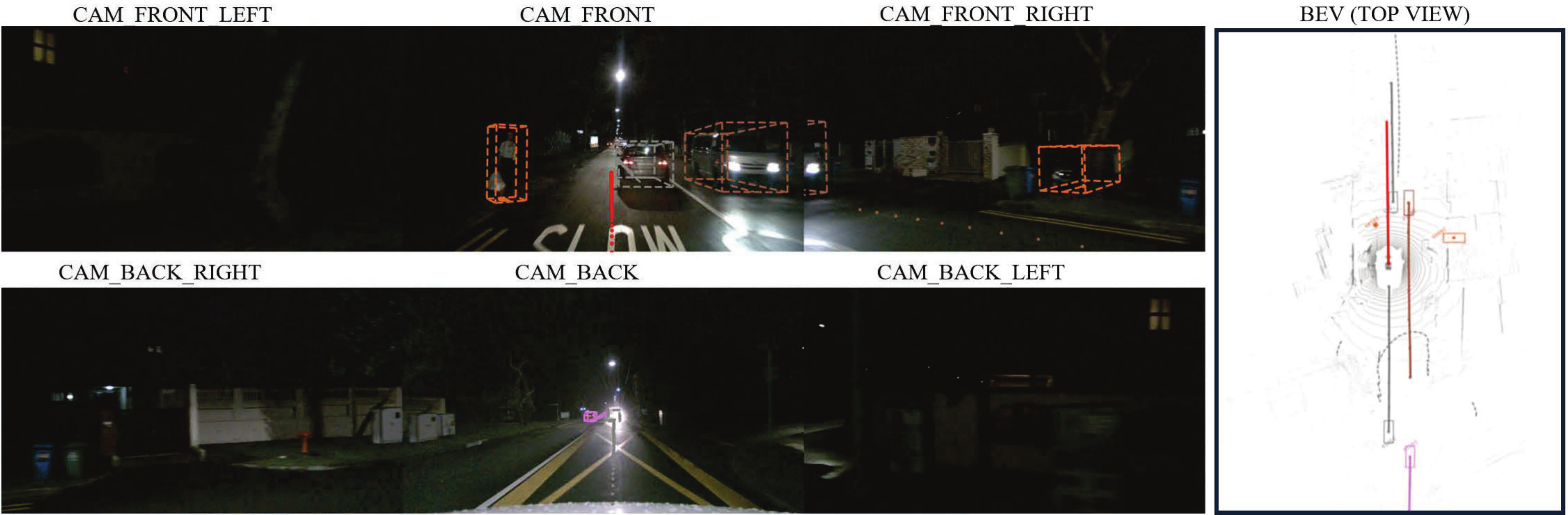}

  \caption{Visualization results in different scenes. \model demonstrates remarkable performance across various scenarios, delivering reasonable planning results. We present the output results of \model under three different scenarios. The first row is the intersection scene, the second row is the turning scene, and the third row is the nighttime scene. The first three columns of each row are visualizations from different camera perspectives. The last column shows the visualization result from the BEV.}
  \label{fig:normal_vis}
  \vspace{-0.25cm}
\end{figure}
\newpage
\begin{figure}[h]
  \centering
  \includegraphics[width=1\linewidth]{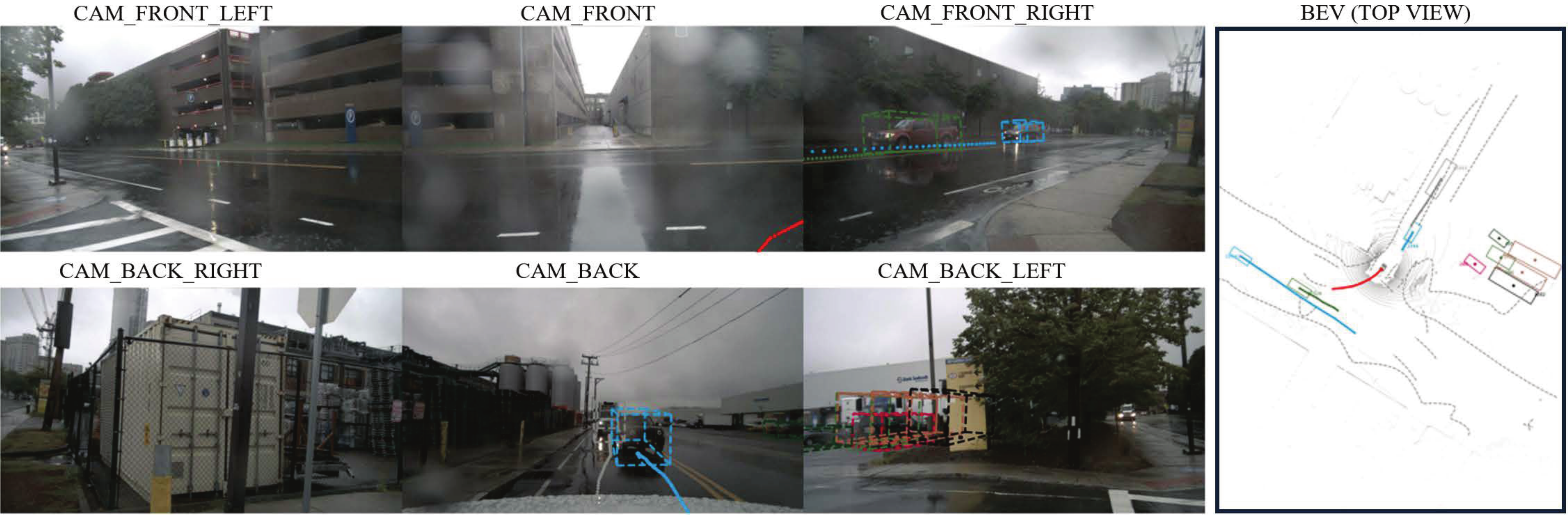}
      \\
    ~
    \\
   \includegraphics[width=1\linewidth]{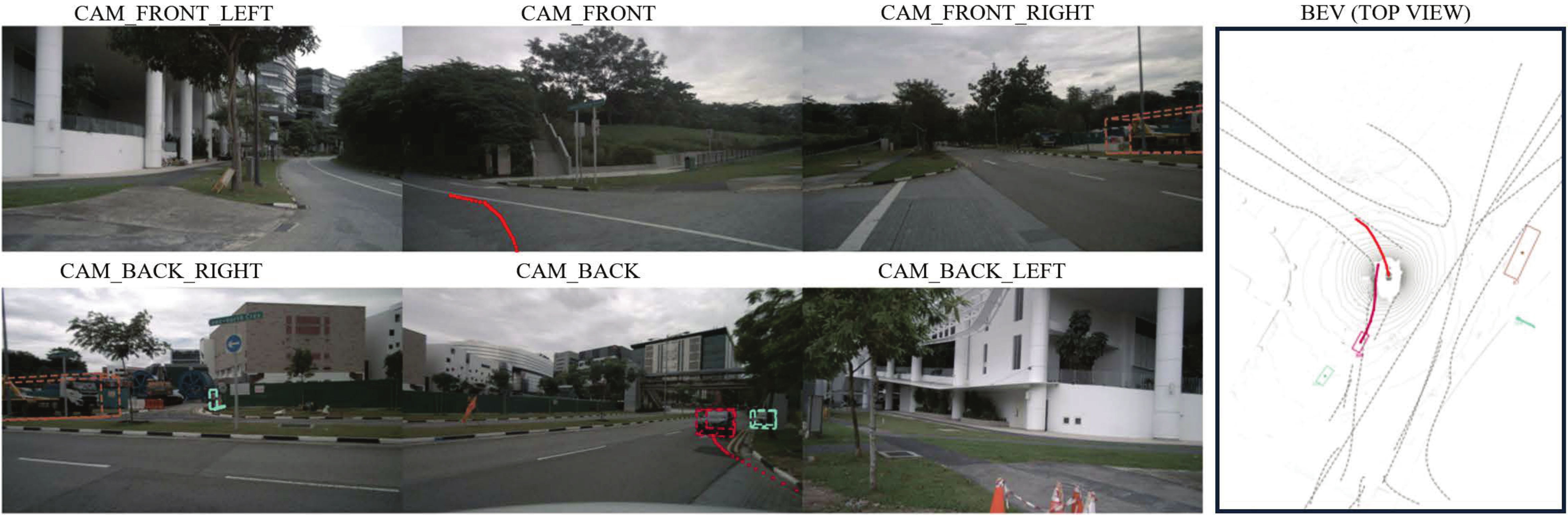}
       \\
    ~
    \\
  \includegraphics[width=1\linewidth]{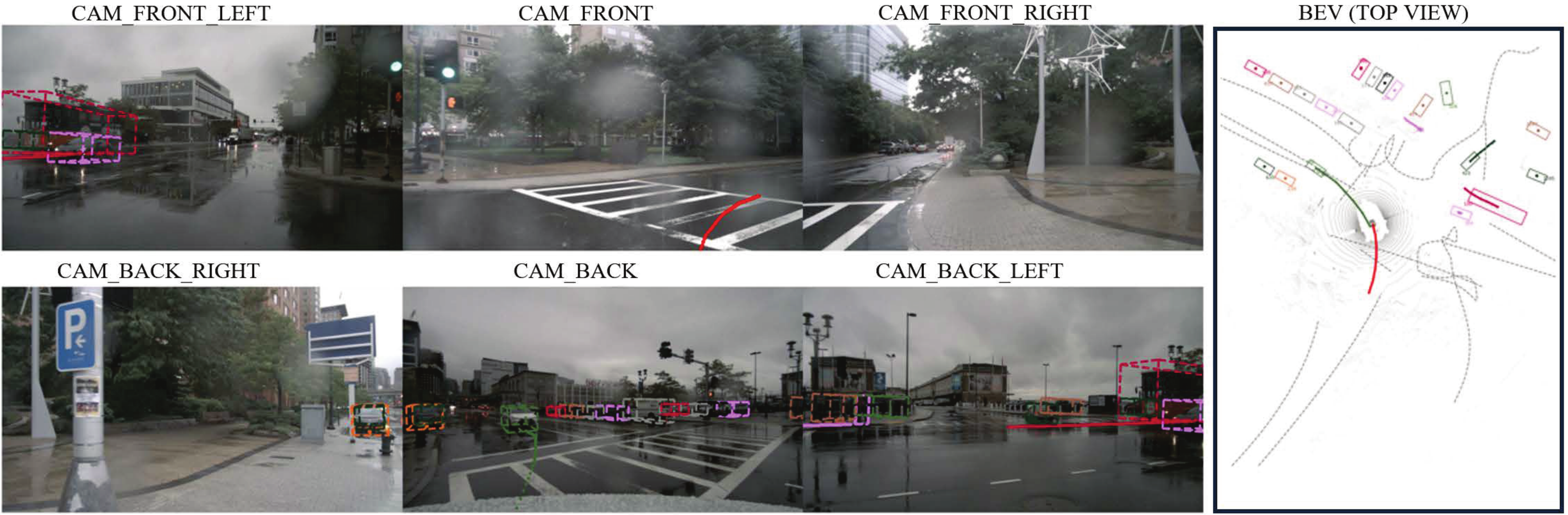}

\caption{Visualization of Intersection Decision-Making Scenarios. On the nuScenes dataset, straight-going scenarios make up the vast majority of the \texttt{train} set. However, as shown SparseAD accurately predicts the future trajectories of other vehicles, and under the input of commands, successfully completes the intersection decision-making task, which proves the the ability of SparseAD to make correct decisions in intersection decision-making scenarios.}
  \label{fig:turn_vis}
\end{figure}
\newpage
\begin{figure}[h]
    \centering
    \includegraphics[width=1\linewidth]{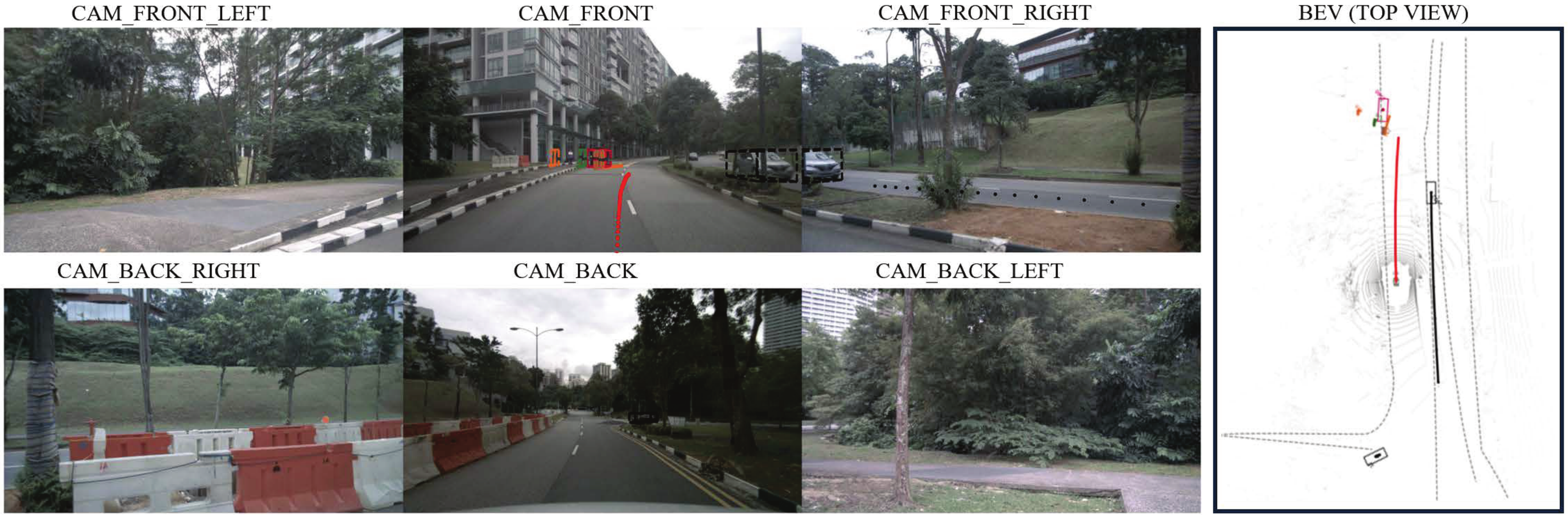}
    \\
    ~
    \\
    \includegraphics[width=1\linewidth]{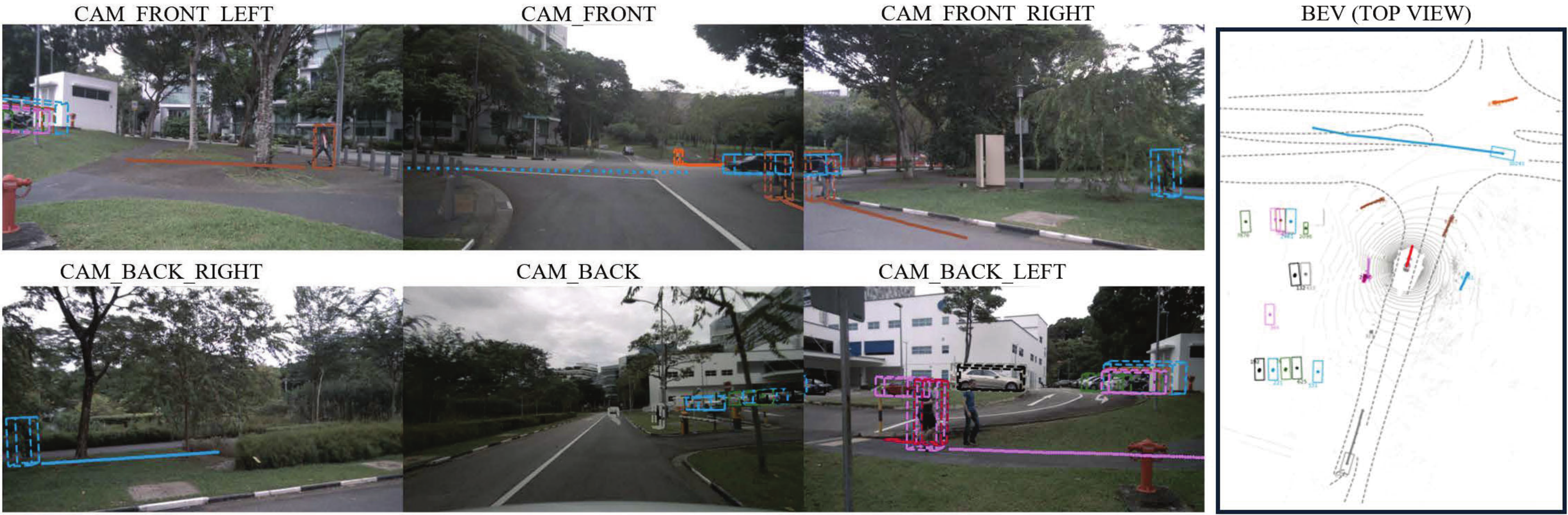}
    \\
    ~
    \\
    \includegraphics[width=1\linewidth]{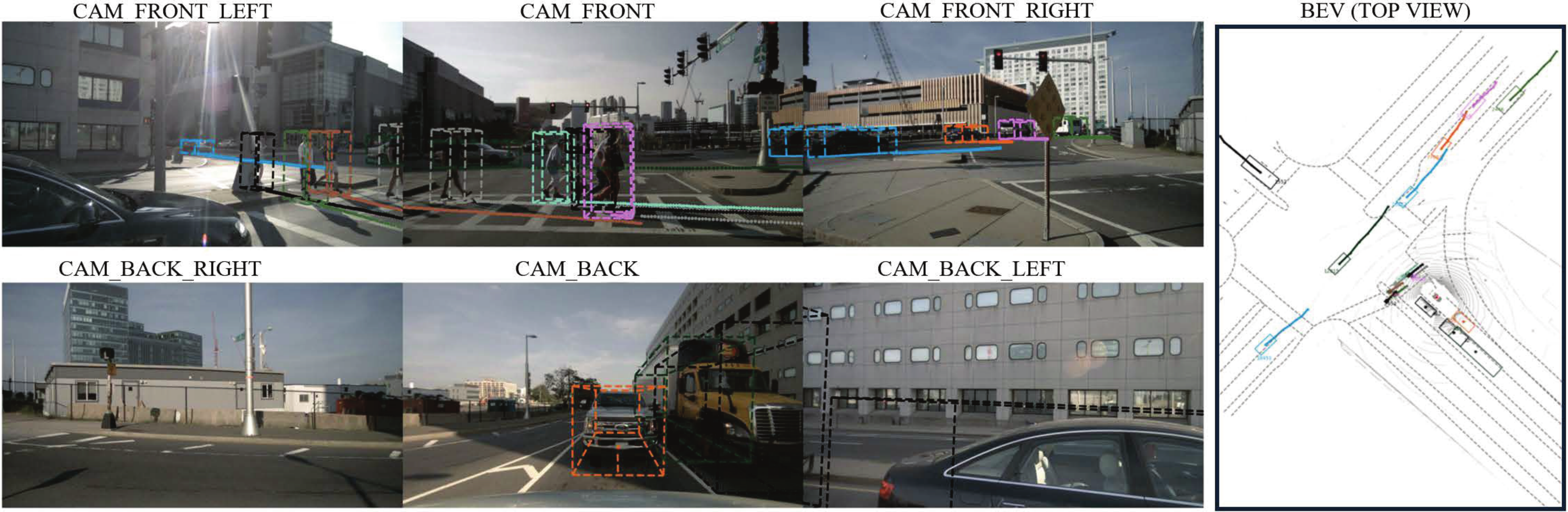}

    \caption{Visualization of Obstacle Avoidance Scenarios. In the first scenario, \model noticed pedestrians in the distance and adjusted the trajectory direction from afar to avoid a collision. In the latter two scenarios, \model detected pedestrians and vehicles crossing laterally at intersections and made decisions to decelerate and stop to wait respectively.}
    \label{fig:avoid_vis}
  \end{figure}
\begin{figure}[h]
    \centering
    \includegraphics[width=1\linewidth]{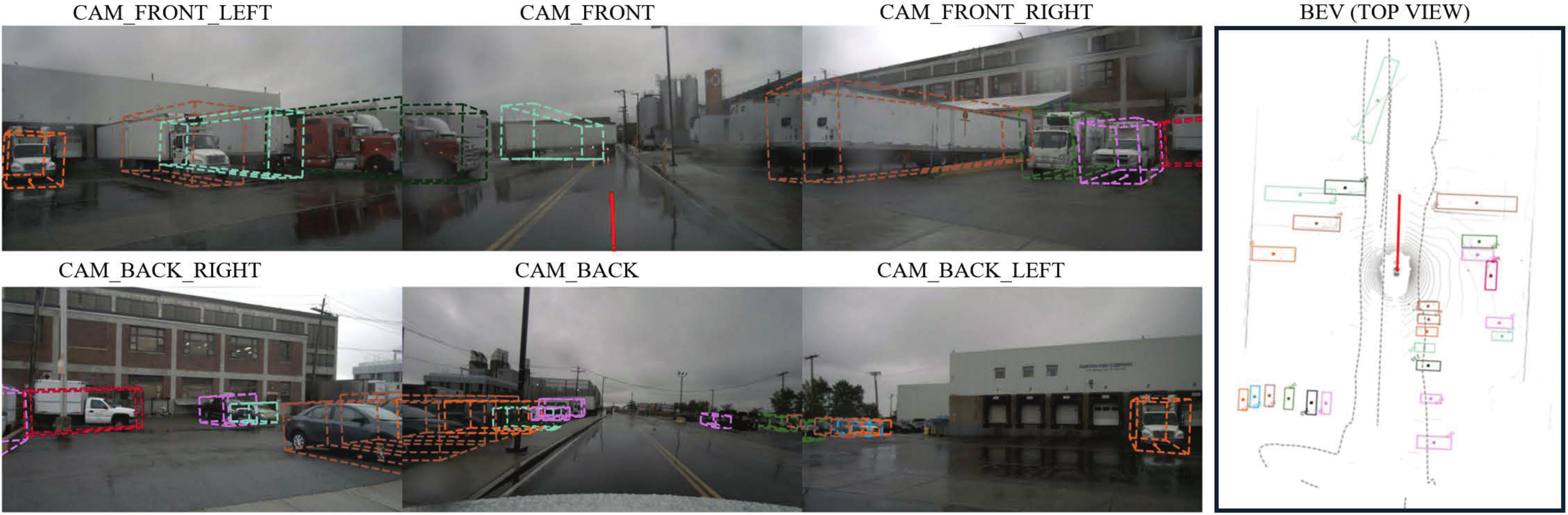}
            \\
    ~
    \\
    \includegraphics[width=1\linewidth]{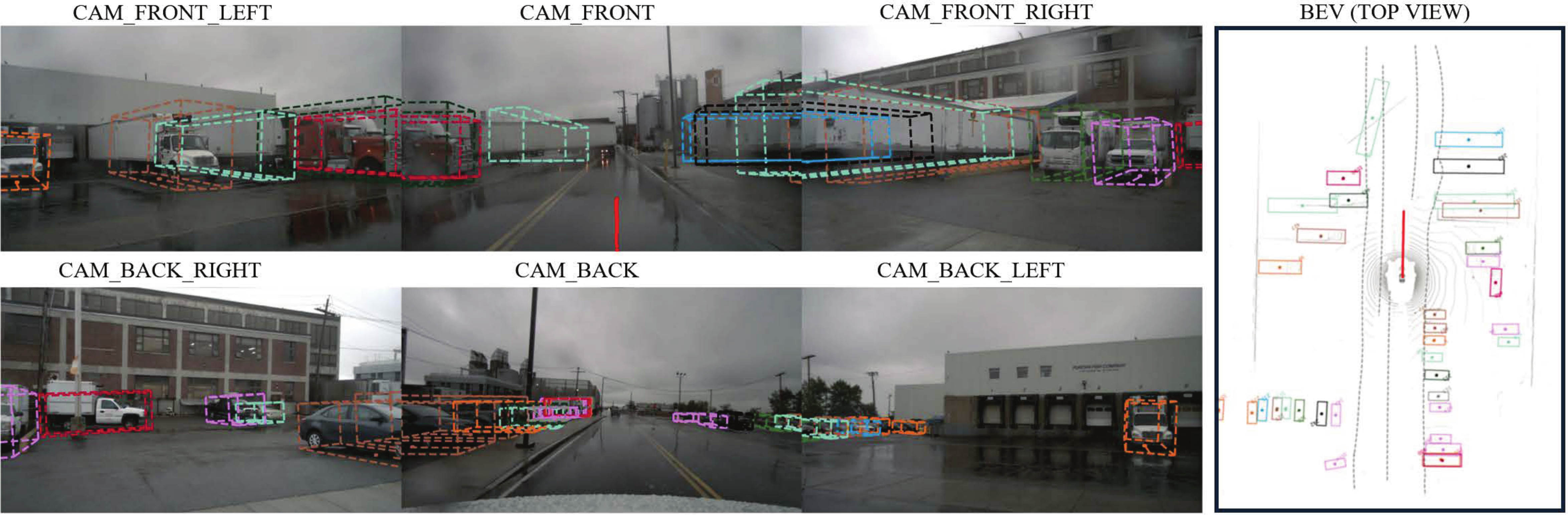}
            \\
    ~
    \\
    \includegraphics[width=1\linewidth]{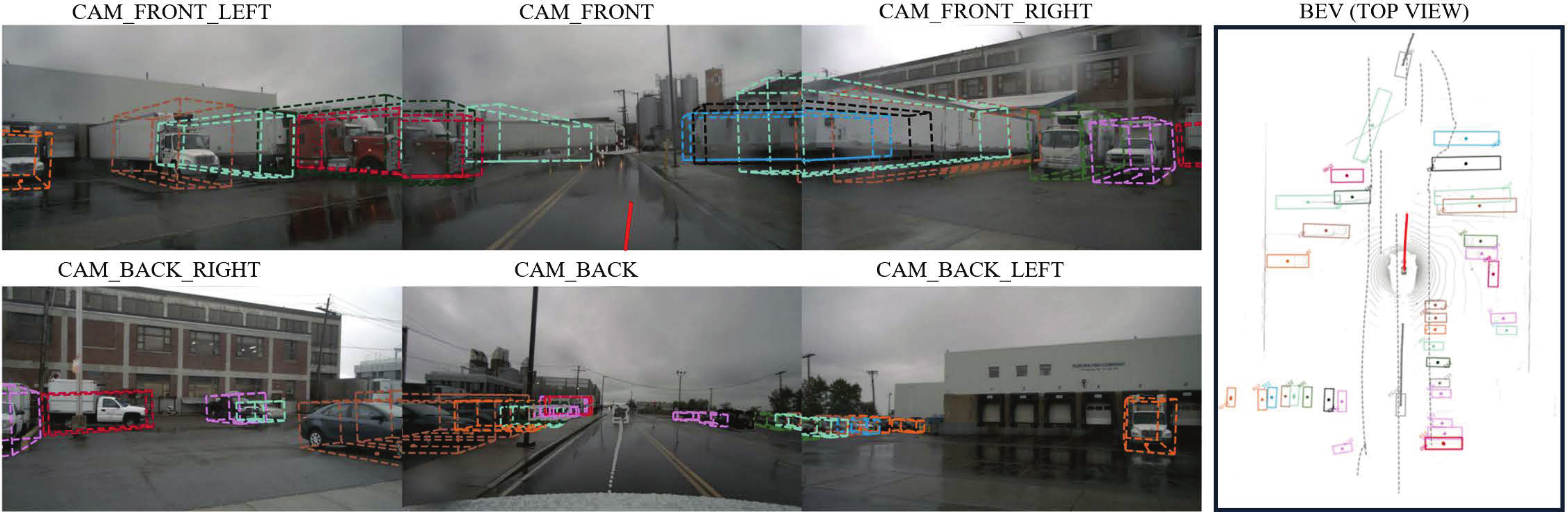}

    \caption{Failure Case. The three rows represent the output results of \model at consecutive frames. As can be seen, the ego-vehicle remains stationary, yet \model still provides a straight-ahead planning result. However, from the perception results, proceeding straight in this scenario is not an unreasonable choice. This indicates that \model does not fully utilize the previous states of the ego-vehicle, leading to a discrepancy with the ground truth.}
    \label{fig:fail_vis}
  \end{figure}

\end{document}